\documentclass[runningheads]{llncs}

\usepackage{eccv}

\usepackage{eccvabbrv}

\usepackage{graphicx}
\usepackage{booktabs}
\usepackage{amsfonts}
\usepackage{amsmath}
\usepackage{graphicx}
\usepackage{booktabs}
\usepackage{overpic}
\usepackage{wrapfig}
\usepackage{empheq}
\usepackage{comment}
\usepackage{makecell}
\usepackage{multirow}
\usepackage{multicol}
\usepackage{float}
\usepackage{marvosym}
\DeclareMathOperator*{\argmax}{arg\,max} 

\usepackage[accsupp]{axessibility}  

\usepackage{hyperref}

\usepackage{orcidlink}

\begin{document}

\title{M$^3$: Dense Matching Meets Multi-View Foundation Models for Monocular \\ Gaussian Splatting SLAM} 

\titlerunning{Multi-View Matching for Monocular SLAM}





\author{
Kerui Ren\inst{1,2} \enspace
Guanghao Li\inst{3,4} \enspace
Changjian Jiang\inst{2,5} \enspace
Yingxiang Xu\inst{6,2} \enspace
Tao Lu\inst{2} \enspace
Linning Xu\inst{7,2} \enspace
Junting Dong\inst{2} \enspace
Jiangmiao Pang\inst{2} \enspace
Mulin Yu\inst{2}{\Letter} \enspace
Bo Dai\inst{8}{\Letter}
} 

\authorrunning{K. Ren et al.}

\institute{
\vspace{0.5em}\inst{1}Shanghai Jiao Tong University,
\text{ }\inst{2}Shanghai Artificial Intelligence Laboratory,
\newline\inst{3}Fudan University,
\text{ }\inst{4}Shanghai Innovation Institute,
\text{ }\inst{5}Zhejiang University,
\newline\inst{6}Beijing Institute of Technology,
\text{ }\inst{7}The Chinese University of Hong Kong,
\newline\inst{8}The University of Hong Kong
}

\maketitle

\begin{figure*}[htbp]
\includegraphics[width=\linewidth]{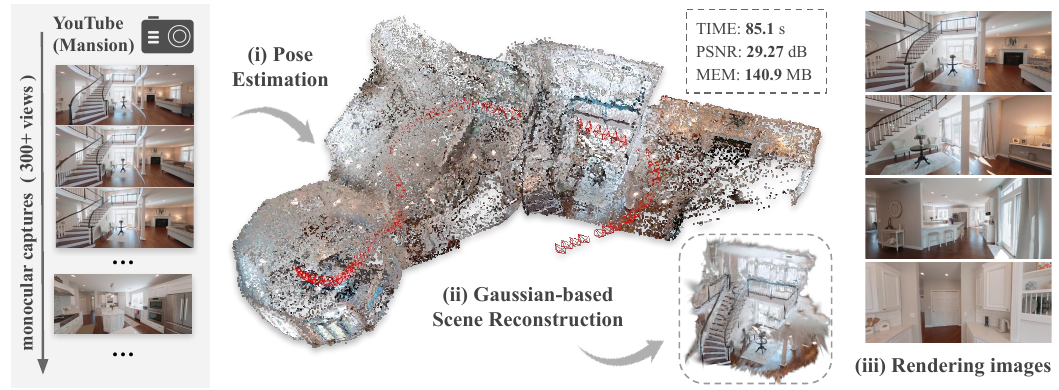}
\caption{
We demonstrate the execution of our M$^3$ pipeline on a challenging manor sequence. Our approach delivers robust, high-precision pose estimation while achieving high-fidelity scene reconstruction from monocular video sequences. Please refer to our project page for more demos: \href{https://city-super.github.io/M3/}{\textcolor{magenta}{\textbf{https://city-super.github.io/M3/}}}
}
\label{fig:teaser}
\end{figure*}

\begin{abstract}
Streaming reconstruction from uncalibrated monocular video remains challenging, as it requires both high-precision pose estimation and computationally efficient online refinement in dynamic environments. While coupling 3D foundation models with SLAM frameworks is a promising paradigm, a critical bottleneck persists: most multi-view foundation models estimate poses in a feed-forward manner, yielding pixel-level correspondences that lack the requisite precision for rigorous geometric optimization. To address this, we present \textbf{M}$^3$, which augments the \textbf{M}ulti-view foundation model with a dedicated \textbf{M}atching head to facilitate fine-grained dense correspondences and integrates it into a robust \textbf{M}onocular Gaussian Splatting SLAM. M$^3$ further enhances tracking stability by incorporating dynamic area suppression and cross-inference intrinsic alignment. Extensive experiments on diverse indoor and outdoor benchmarks demonstrate state-of-the-art accuracy in both pose estimation and scene reconstruction. Notably, M$^3$ reduces ATE RMSE by 64.3\% compared to VGGT-SLAM 2.0 and outperforms ARTDECO by 2.11 dB in PSNR on the ScanNet++ dataset.
  \keywords{SLAM \and Streaming Reconstruction \and 3D Gaussian Splatting}
\end{abstract}

\section{Introduction}
\label{sec:intro}

3D scene reconstruction has become a fundamental capability in computer vision, enabling applications ranging from robotic perception to large-scale scene digitization~\cite{zhou2025hugsim,huang2026soma,yu2026gaussexplorer,yang2025novel}. Recently, the field has been revolutionized by two paradigms: per-scene optimization, such as 3D Gaussian Splatting (3DGS)~\cite{kerbl20233d}, which delivers high-fidelity rendering, and feed-forward geometric foundation models~\cite{leroy2024grounding,wang2025pi,wang2025vggt}, which infer dense priors in a single pass. However, most existing foundation models are inherently batch-oriented, designed to process a fixed set of images jointly. This offline nature precludes real-time feedback and limits scalability in open-ended environments, underscoring the urgent need for streaming reconstruction, where camera trajectories and scene geometry are incrementally updated as new observations arrive.

Existing efforts toward streaming 3D reconstruction generally follow two trajectories, yet both face significant hurdles. The first family attempts to adapt feed-forward models to a streaming context by incorporating memory mechanisms that summarize past observations to predict geometry incrementally~\cite{spann3r,long3r,point3r}. While these methods are efficient, they typically produce low-resolution results and struggle with cumulative drift, as they lack the iterative global refinement mechanisms found in classical SLAM. The second family instead integrates foundation-model priors into a SLAM pipeline to guide optimization~\cite{murai2024_mast3rslam,artdeco,maggio2025vggt}. However, these approaches are often trapped in a fundamental trade-off: pairwise-prior methods, such as MASt3R-SLAM~\cite{murai2024_mast3rslam}, suffer from redundant computation and quadratic complexity, whereas multi-frame prior methods like VGGT-SLAM 2.0~\cite{maggio2026vggt} provide global geometry but lack the pixel-level dense correspondences necessary for rigorous geometric optimization.

We argue that the primary bottleneck in current multi-view foundation models is their disproportionate focus on individual scene geometry at the expense of inter-view relational consistency. While these models produce impressive 3D structures, they are often blind to the precise pixel-to-pixel associations across frames. Without such fine-grained correspondences, the SLAM backend cannot establish the strong epipolar constraints required for Bundle Adjustment (BA), leading to catastrophic failures like ghosting artifacts or trajectory divergence in complex sequences. Consequently, fine-tuning foundation models specifically to recover dense matching is no longer an option, but a necessity to unlock their full potential for downstream SLAM tasks.

To bridge this gap, we propose M$^3$, a streaming 3D reconstruction framework that tightly couples a multi-view foundation model with a robust SLAM pipeline. Our approach first enhances a state-of-the-art multi-view geometric foundation model by introducing a dedicated dense matching head, specifically trained to recover pixel-level correspondences. This enables the SLAM framework to leverage the foundation model’s geometry for accurate, high-frequency pose refinement. Unlike previous black-box integrations, M$^3$ performs a single feed-forward inference over both historical keyframes and incoming frames to simultaneously update geometry and tracking, significantly reducing redundant model invocations. Furthermore, we introduce a dynamic region identification module to detect and suppress transient objects, ensuring stable static scene reconstruction in real-world environments. Extensive experiments across diverse indoor and outdoor benchmarks demonstrate that M$^3$ achieves state-of-the-art accuracy in both pose estimation and 3D reconstruction, while maintaining competitive efficiency on long-duration monocular video streams.

In summary, our core contributions are as follows:

\begin{itemize}
    \item We introduce a dedicated matching head to a multi-view foundation model, leveraging pixel-level descriptors to facilitate refined cross-frame dense matching for rigorous geometric optimization.
    \item We propose M$^3$, a SLAM framework leveraging a multi-view foundation model to simultaneously facilitate frontend tracking and backend global optimization via a single feed-forward inference.
    \item Extensive experiments across diverse benchmarks demonstrate that M$^3$ delivers state-of-the-art accuracy in both pose estimation and 3D reconstruction, while maintaining high computational efficiency on long-duration monocular sequences.
\end{itemize}

\section{Related work}
\label{sec:related_work}

\subsection{Learning-based and Foundation Model-based SLAM}
In recent years, visual Simultaneous Localization and Mapping (SLAM) has evolved from hand-crafted feature-based pipelines~\cite{campos2021orb,forster2014svo,engel2017direct} toward end-to-end learning-based frameworks~\cite{teed2021droid, teed2023deep}. DROID-SLAM~\cite{teed2021droid} established a strong benchmark by integrating a Gated Recurrent Unit (GRU) with a differentiable bundle adjustment (BA) layer, enabling iterative refinement of camera poses and pixel-wise disparities.
Recent SLAM frameworks increasingly incorporate geometric priors or foundation models to improve robustness and accuracy. For example, MegaSaM~\cite{li2025megasam} addresses the challenges posed by in-the-wild videos by integrating monocular depth priors with motion probability networks, enabling robust handling of dynamic objects and low-parallax motion. Moving toward more generalizable architectures, MASt3R-SLAM~\cite{murai2024_mast3rslam} leverages dense pointmap regression for calibration-free tracking, allowing the framework to operate under unknown camera parameters. To resolve the projective ambiguity inherent in such uncalibrated sequences, VGGT-SLAM~\cite{maggio2025vggt} formulates the optimization problem on the $\mathbf{SL}(4)$ manifold via factor graph optimization, thereby ensuring global consistency across submaps while accounting for 15-DOF projective transformations. 

\subsection{3D Scene Reconstruction}
The development of 3D scene reconstruction has progressed from classical geometric formulations, such as Structure-from-Motion (SfM) \cite{schonberger2016structure} and Multi-View Stereo (MVS)~\cite{schonberger2016pixelwise}, toward differentiable neural representations. Early implicit approaches, exemplified by Neural Radiance Fields (NeRF)~\cite{mildenhall2021nerf}, demonstrated remarkable fidelity in novel view synthesis; however, their reliance on computationally intensive ray marching limited their applicability in real-time scenarios. This paradigm shifted with the introduction of 3D Gaussian Splatting (3DGS)~\cite{kerbl20233d}, which represents scenes using explicit anisotropic Gaussian primitives. By leveraging tile-based differentiable rasterization, 3DGS enables real-time rendering while supporting explicit optimization~\cite{lu2024scaffold,jiang2025horizon,ren2024octree}.

Recent works have extended 3DGS to support a broader range of tasks. For instance, several studies have refined these representations by introducing 2D Gaussian primitives \cite{huang20242d, dai2024high} to improve surface accuracy, as well as incorporating depth and normal consistency, to mitigate rendering artifacts. Consequently, these point-based representations not only achieve photorealistic rendering quality but also maintain geometric accuracy for complex spatial and temporal tasks.
Moreover, 4D Gaussian Splatting~\cite{wu20244d} and its variants~\cite{lin2024gaussian,li2024spacetime,zhu2024motiongs} incorporate temporal modeling by associating Gaussians with time-dependent deformation fields or velocity vectors. These approaches enable high-fidelity reconstruction of non-rigid motions and transient scene elements while preserving the real-time rendering advantages of the original splatting framework. 

\subsection{Streaming Reconstruction}
Streaming reconstruction aims to incrementally build 3D models from sequential sensor data streams while maintaining low latency~\cite{wang2026towards}. Recent methods extend this paradigm by combining online pose estimation with 3DGS to enable real-time streaming reconstruction. GS-SLAM~\cite{yan2024gs} was among the first to integrate 3DGS into a dense SLAM framework, employing adaptive Gaussian expansion and coarse-to-fine tracking to jointly address mapping and rendering. To enable streaming reconstruction in large-scale environments, Onthefly-NVS~\cite{meuleman2025fly} introduced a pixel spawning mechanism along with a sliding-window anchor strategy, allowing real-time processing of kilometer-scale scenes. For dynamic scenes, Instant4D~\cite{luo2025instant4d} demonstrated minute-level 4D reconstruction from monocular video by employing simplified isotropic Gaussians, reducing redundancy by up to 90\%. 

More recent works further enhance robustness by incorporating priors from geometric or vision foundation models~\cite{artdeco, jiang2026planing, cheng2025outdoor, zhangflash}. 
For example, ARTDECO~\cite{artdeco} combines MASt3R-based feed-forward predictions with a level-of-detail representation to maintain global consistency under unconstrained streaming inputs. 
To address the limited surface coherence of pure Gaussian-based representations, PLANING~\cite{jiang2026planing} employs a hybrid triangle-Gaussian representation within the streaming reconstruction framework. 
By decoupling stable geometric anchors from neural appearance modeling, this design enables structured surface reconstruction suitable for downstream tasks.
\section{Method}
\label{sec:method}

\begin{figure}[t!]
\centering
\includegraphics[width=\linewidth]{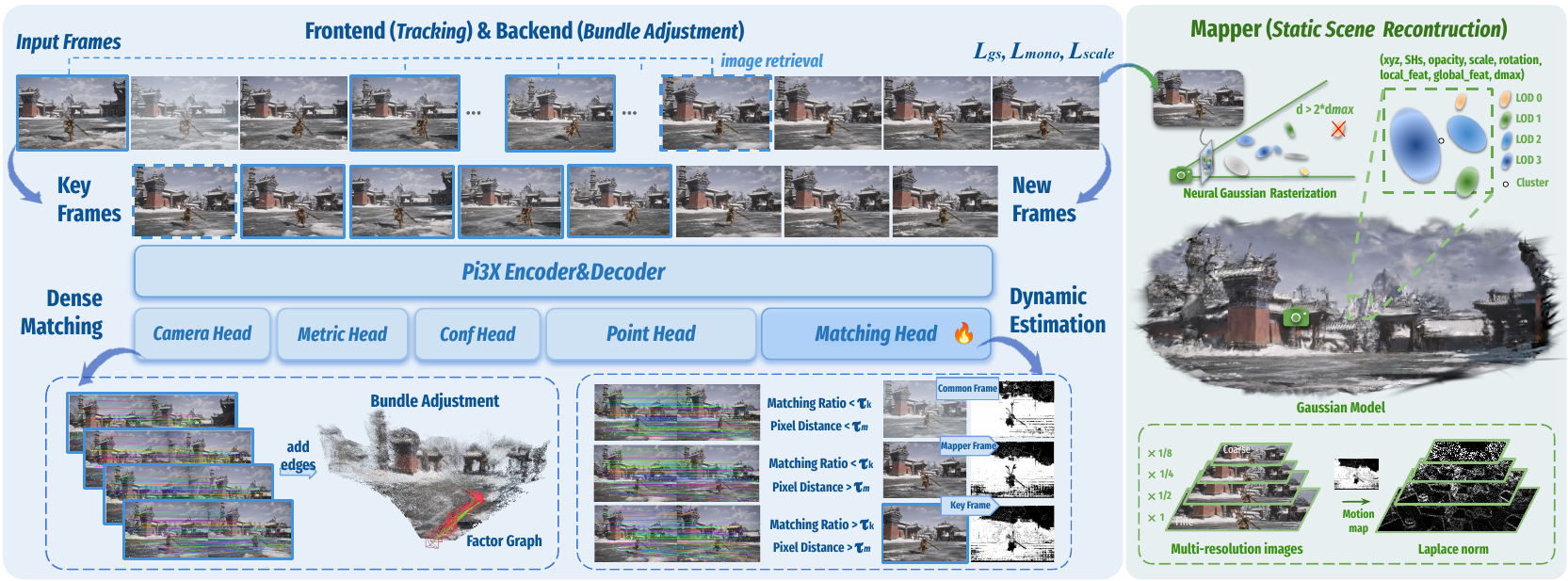}
\caption{
\textbf{The M$^3$ Pipeline.} The framework consists of joint tracking and global optimization for pose estimation and a mapper for scene reconstruction. For monocular sequences, Pi3X processes retrieved historical keyframes and new frames in one inference to facilitate factor graph construction and keyframe selection. Following the Neural Gaussian and LOD architecture of ARTDECO~\cite{artdeco}, Gaussians are initialized via Laplacian norm and optimized jointly with camera poses.
}
\label{fig:method}
\end{figure}

Fig.~\ref{fig:method} illustrates the overall pipeline of M$^3$, an efficient streaming framework for scene reconstruction from uncalibrated monocular videos $\{I_i\}_{i=1}^{N}$. Specifically, our method jointly estimates camera intrinsics $\mathbf{K} \in \mathbb{R}^{3 \times 3}$ and camera poses $\{\mathbf{R}_i, \mathbf{t}_i\}_{i=1}^{N}$, while reconstructing a set of neural Gaussians $\{G_j\}_{j=1}^{M}$ representing the underlying static 3D scene. Recent works~\cite{maggio2025vggt, maggio2026vggt, murai2024_mast3rslam, artdeco} have attempted to improve SLAM efficiency, accuracy, and robustness by integrating foundation models~\cite{leroy2024grounding, wang2025vggt} into SLAM pipelines. However, these approaches either suffer from computational redundancy due to repeated pairwise model inferences or lack sufficient geometric precision because pixel-level correspondences are not explicitly established. To address these limitations, we propose M$^3$, which incorporates a variant of $\pi^3$~\cite{wang2025pi}, Pi3X, augmented with dense pixel-level matching, as detailed in Sec.~\ref{sec:densematching}. In addition, we explicitly filter dynamic transient regions to better adapt to complex real-world environments. We further integrate the foundation model into a unified and tightly coupled frontend--backend SLAM framework, as illustrated in Fig.~\ref{fig:method}. For notational brevity, we denote by $\mathbf{X}_i^j$ the point map of the $i$-th frame transformed into the coordinate frame of the $j$-th frame. In particular, $\mathbf{X}_i \triangleq \mathbf{X}_i^i$ denotes the point map in its own coordinate.

\subsection{Dense Matching through Foundation Model}
\label{sec:densematching}


As demonstrated in~\cite{murai2024_mast3rslam, artdeco}, per-pixel dense matching is crucial for tightly integrating foundation models into SLAM frameworks. However, existing foundation models with dense matching capability~\cite{leroy2024grounding} operate primarily in a pairwise manner, processing only two images per inference. When extended to multi-view sequences, this design leads to substantial redundant computations. To address this limitation, we augment Pi3X with a dense matching module. Originally, Pi3X is designed for efficient camera pose and depth estimation from arbitrary video frames. For each input frame, it predicts a local point map $\mathbf{X}_i \in \mathbb{R}^{H \times W \times 3}$, a camera-to-world transformation $\mathbf{T}_i \in \mathbb{R}^{4 \times 4}$, and a geometric confidence map $\mathbf{C}_i \in \mathbb{R}^{H \times W \times 1}$. Compared to its previous version~\cite{wang2025pi}, the enhanced model produces smoother reconstructions and supports approximate metric scale, thereby providing a stronger geometric prior for the downstream SLAM framework. In the following, we describe the model architecture, training strategy, dense matching, and dynamic object handling.

\subsubsection{Model Architecture.} 
We extend the Pi3X architecture by incorporating a matching head inspired by MASt3R~\cite{leroy2024grounding} to concurrently predict dense feature descriptors $\mathbf{D} \in \mathbb{R}^{N \times H \times W \times d}$ and matching confidence maps $\mathbf{Q} \in \mathbb{R}^{N \times H \times W \times 1}$, where $d$ denotes the dimensionality of the descriptor, 24 by default.
The matching head consists of a Dense Prediction Transformer block ($\mathrm{DPT_{desc}}$) and a 2-layer $\mathrm{MLP_{desc}}$ interleaved with a non-linear GELU activation function~\cite{hendrycks2016gaussian}. To ensure matching stability, each local feature is normalized to unit norm. The output of the matching head is formulated as:
\begin{equation}
(\mathbf{D}, \mathbf{Q}) = \mathrm{MLP_{desc}}(\mathrm{DPT_{desc}}(\mathbf{H})),
\end{equation}
where $\mathbf{H}$ represent the intermediate feature representations extracted from the Pi3X decoder. More architectural details can be found in the supplementary.

\subsubsection{Loss function and training.}

To preserve the geometric priors and metric scale of the foundation model, we freeze the encoder, decoder, and existing heads, and fine-tune only the matching head. 
The $\mathrm{DPT}_{\text{desc}}$ module is initialized with the pre-trained parameters of the point head, exploiting the shared structural knowledge between geometric prediction and dense matching to accelerate convergence. Within a batch of $N$ images, we treat $I_1$ as the reference and establish ground-truth correspondences $\hat{\mathcal{M}}_k$ between $I_1$ and each subsequent frame $I_k$ ($k=2,\dots,N$) by identifying pixel pairs with coincident 3D coordinates:
$
\hat{\mathcal{M}}_k \;=\; \left\{ (u,v)\ \middle|\ \hat{\mathbf{X}}_{1,u}^{1} =\hat{\mathbf{X}}_{k,v}^{1}\right\},
$
where $u$ and $v$ denote pixel indices in $I_1$ and $I_k$, respectively. The overall training objective is a weighted combination of the matching loss and a confidence regularization term:
\begin{equation}
\mathcal{L}
\;=\;
-\frac{1}{N-1}\sum_{k=2}^{N}\Big( \mathbf{Q}_k\,\mathcal{L}^{k}_{\text{match}}-\alpha \log \mathbf{Q}_k \Big),
\end{equation}
where $\alpha$ is a balancing hyperparameter. For each image pair, $\mathcal{L}^{k}_{\text{match}}$ adopts a symmetric InfoNCE objective with bidirectional terms to encourage mutual descriptor consistency:
\begin{equation}
\mathcal{L}^{k}_{\text{match}}
\;=\;
-\sum_{(u,v)\in \hat{\mathcal{M}}_k}
\left(
\log \frac{s_{\tau}(u,v)}{\sum_{w \in \mathcal{P}_{1}} s_{\tau}(w,v)}
\;+\;
\log \frac{s_{\tau}(u,v)}{\sum_{w \in \mathcal{P}_{k}} s_{\tau}(u,w)}
\right),
\end{equation}
where the similarity score is computed as
$
s_{\tau}(u,v) \;=\; \exp\!\Big(-\tau\,\mathbf{D}_{1,u}^{\top}\mathbf{D}_{k,v}\Big).
$ Here, $\tau$ denotes the temperature hyperparameter, and $\mathbf{D}_1$ and $\mathbf{D}_k$ are the predicted dense feature descriptors of images $I_1$ and $I_k$, respectively. The sets $\mathcal{P}_1$ and $\mathcal{P}_k$ denote the pixel sets in $I_1$ and $I_k$, respectively, that form the correspondence pairs in $\hat{\mathcal{M}}_k$.


\subsubsection{Dense matching for SLAM.} After incorporating the matching head into Pi3X, we describe how pixel-wise correspondences across frames are computed for SLAM. Given $N$ input images $\{I_i\}_{i=1}^{N}$, the enhanced Pi3X outputs per-frame estimates $\{\mathbf{X}_i, \mathbf{T}_i, \mathbf{C}_i, \mathbf{D}_i, \mathbf{Q}_i\}$. For an image pair $(I_i, I_j)$, we define the pixel correspondence from $I_i$ to $I_j$ as $\mathcal{M}(I_i, I_j)$. 
Using the predicted poses, we initialize matching by transforming $\mathbf{X}_i$ into the coordinate frame of $I_j$, given by
$
\mathbf{X}_i^j = \mathbf{T}_j^{-1} \mathbf{T}_i \mathbf{X}_i
$,
where $\mathbf{X}_i^j$ represents the points of frame $i$ expressed in the local coordinate framework of $I_j$. The pose-guided transformation provides a geometry-aware initialization that restricts the correspondence search to a small spatial region, eliminating the need for exhaustive global matching. 

We then refine the correspondence by searching within a local neighborhood of radius $r$ around the projected location and selecting the pixel $\mathbf{p}^*$ that maximizes descriptor similarity:
\begin{equation}
\mathbf{p}^* = 
\argmax_{\|\mathbf{p} - \phi(\mathbf{X}_i^j(\mathbf{q}))\| \le r}
\langle \mathbf{D}_i(\mathbf{q}), \mathbf{D}_j(\mathbf{p}) \rangle,
\label{eq:fine_match}
\end{equation}
where $\mathbf{q}$ denotes a pixel in $I_i$, 
$\phi(\cdot)$ is the projection function from 3D coordinates to the image plane of $I_j$, 
and $\langle \cdot, \cdot \rangle$ denotes cosine similarity. 

By constraining matching to a pose-guided local neigborhood, the computational cost is reduced from quadratic global search to linear local refinement, while preserving high matching accuracy. This coarse-to-fine scheme enables efficient dense correspondence estimation for the downstream SLAM pipeline.

{\subsubsection{Dynamic Region Estimation.} Dynamic objects in real-world videos may introduce geometric artifacts and degrade pose estimation accuracy. To enhance robustness, we introduce a descriptor-based motion estimation module that predicts a motion map $\mathbf{M}_i \in [0,1]^{H \times W \times 1}$ to suppress dynamic regions. Given a reference keyframe $k$, we project its descriptor map $\mathbf{D}_k$ into the current frame $i$ using the estimated transformation, obtaining the warped descriptor map $\mathbf{D}_k^i$. In static regions, the warped descriptors $\mathbf{D}_k^i$ should align well with the predicted descriptors $\mathbf{D}_i$, resulting in high feature similarity. 
In contrast, dynamic objects or occlusions produce significant descriptor discrepancies. To maintain temporal consistency, we further modulate the similarity using the motion map of the reference frame. The motion map of frame $I_i$ is computed as:
\begin{equation}
\mathbf{M}_i = \langle \mathbf{D}_k^i, \mathbf{D}^i_i \rangle \odot \mathbf{M}_k^i,
\end{equation}
where $\langle \cdot, \cdot \rangle$ denotes pixel-wise dot product and $\mathbf{M}_k^i$ is the motion map of the last keyframe warped into frame $i$. Pixels with low consistency are down-weighted during optimization, reducing trajectory drift and reconstruction artifacts.

\subsection{M$^3$ Framework}
\label{sec:framework}


%
Conventional SLAM frameworks typically decouple frontend tracking from backend global optimization. As a result, the foundation model is invoked separately for frontend tracking and multiple times for backend global bundle adjustment, leading to redundant computation and potentially unstable tracking. In contrast, we leverage the multi-view processing capability of the enhanced Pi3X, which supports up to 16 frames in a single inference pass, to tightly couple the frontend and backend modules. In this section, we first describe how the enhanced Pi3X is incorporated into the SLAM framework. We further describe the optimization of camera parameters and the reconstruction of the Gaussian model in M$^3$.

\subsubsection{Sliding Window Management.} To streamly process the incoming video, we maintain a sliding window of length $L$, which is partitioned into $K$ slots for historical keyframes and $L-K$ for incoming frames. Specifically, the $K$ historical keyframes contain the last keyframe $I_k$ and the else $K-1$ most relevant keyframes,  where the $K-1$ relevant keyframes retrieved via SALAD~\cite{izquierdo2024optimal} descriptors following the strategies in VGGT-SLAM~\cite{maggio2025vggt} and VGGT-Long~\cite{deng2025vggt}. Specifically, if the retrieved keyframes are temporally distant, a Pi3X-aided loop closure is triggered. By jointly feeding historical keyframes and incoming frames into the model, a single forward pass yields dual-purpose outputs: point maps and descriptors from historical frames are used to update the global factor graph, while predictions for new frames provide the initial poses, point maps, and descriptors required for real-time tracking and keyframe selection. The geometric priors of incoming frames attend to multiple keyframes through cross-attention, which enhances tracking stability by enforcing multi-view geometric consistency. Following the design of~\cite{artdeco}, we categorize incoming frames into \emph{keyframes}, \emph{mapper frames}, and \emph{common frames}. 
Keyframes are used for pose estimation; keyframes and mapper frames are employed for 3D model initialization; and all frames contribute to the optimization of the reconstructed 3D model. A frame is identified as a keyframe if its correspondence ratio with the most recent keyframe $I_k$ falls below a threshold $\tau_k$. 
For mapper frame selection, we compute the pixel displacement between the current frame and the latest keyframe; if the 70th percentile of the displacement exceeds a threshold $\tau_m$, the frame is promoted to a mapper frame. Frames that satisfy neither condition are treated as common frames. More details are provided in the supplementary material.}

\subsubsection{Intrinsic Consistency.}
For each inference, an intrinsic matrix $\mathbf{K}$ can be estimated via RANSAC using the predicted poses and point maps. However, the estimated intrinsics may vary slightly across different inferences due to the inherent scale ambiguity in Pi3X, which leads to inconsistent dense correspondences during streaming processing. To address this issue, we use the intrinsic estimated from the first inference as a reference and align the intrinsics of subsequent inferences accordingly. Specifically, the reference intrinsic $\mathbf{K}_r$ is obtained via RANSAC from the poses and point maps predicted in the first inference. For each subsequent inference $i$, we estimate its intrinsic $\mathbf{K}_i$ in the same manner and align it to $\mathbf{K}_r$. The corresponding point map is then rescaled according to the focal length ratio between $\mathbf{K}_i$ and $\mathbf{K}_r$. This alignment maintains consistent geometry across frames in the sliding window, enabling robust data association.

\subsubsection{Tracking and Global Optimization.}After passing the enhanced Pi3X, we first track them for initialized poses before the global BA optimization. To ensure scale consistency between different inferences, we represent camera poses $\mathbf{T}$ within the $\mathbf{Sim}(3)$ group, which also optimizes scale $s \in \mathbb{R}$ besides rotation $\mathbf{R} \in \mathbf{SO}(3)$ and translation $\mathbf{t} \in \mathbb{R}^3$, given by
$
\mathbf{T} = 
\begin{bmatrix}
s \mathbf{R} \quad& \mathbf{t} \\
\mathbf{0}^\top \quad& 1
\end{bmatrix}.
$
Updates are performed on the Lie algebra $\boldsymbol{\tau} \in \mathfrak{sim}(3)$ via a left-plus operator,
$
\mathbf{T} \leftarrow \boldsymbol{\tau} \oplus \mathbf{T} \triangleq \exp(\boldsymbol{\tau}) \circ \mathbf{T}.
$

Building upon this manifold, we track each incoming frame following the previous framework, adopting the pixel-based optimization from MASt3R-SLAM~\cite{murai2024_mast3rslam}. Specifically, the relative frame pose $\mathbf{T}_{kf} = \mathbf{T}_k^{-1} \mathbf{T}_f$ is estimated by minimizing the reprojection error:
\begin{equation}
E_{t} = \sum_{(m,n) \in \mathcal{M}(I_f, I_k)} \mathbf{M}_k
\left\| 
\frac{\mathbf{p}_{k,n} - \phi \left( \mathbf{T}_{kf}\mathbf{X}^f_{f,m} \right)}
{w(q_{m,n})}
\right\|_\rho ,
\end{equation}
where $\mathbf{p}_{k,n}$ denotes the coordinate of pixel $n$ in $I_k$, 
$q_{m,n} = \sqrt{\mathbf{Q}_{f,m} \mathbf{Q}_{k,n}}$, and 
$\mathcal{M}(I_f, I_k)$ denotes the dense correspondences established in Sec.~\ref{sec:densematching}. 
$\|\cdot\|_\rho$ denotes the Huber kernel, and $w(\cdot)$ is a per-match weighting function following~\cite{murai2024_mast3rslam}.

After tracking each image pose, we perform a global optimization of all keyframe poses by minimizing the collective reprojection error according to the factor graph maintained by our framework:
\begin{equation}
E_{g} = \sum_{(i,j) \in \mathcal{E}} \sum_{(m, n) \in \mathcal{M}(I_j, I_i)} \mathbf{M}_i \left\| \frac{\mathbf{p}_{i,n} - \phi \left( \mathbf{T}_{ij}\tilde{\mathbf{X}}^j_{j,m} \right)}{w(q_{m,n})} \right\|_\rho,
\end{equation}
where $\mathcal{E}$ is the set of edges in the factor graph, and $\tilde{\mathbf{X}}^j_{j}$ denotes the point map maintained by keyframe $j$, which is updated using a weighted average filter~\cite{murai2024_mast3rslam}.

\subsubsection{Neural Gaussian Reconstruction.} Each 3D Gaussian primitive is parameterized by its spatial center $\boldsymbol{\mu} \in \mathbb{R}^3$, rotation $\mathbf{r} \in \mathbf{SO}(3)$, scale $\mathbf{s} \in \mathbb{R}^3$, opacity $\alpha \in [0, 1]$, spherical harmonics (SH) coefficients and $\mathbf{d}_{max}$ for LoD rendering. To promote global consistency while preserving local distinctiveness, we further incorporate an individual feature $\mathbf{f}_l$ and a region-wise feature $\mathbf{f}_g$ that encodes local voxel context. Inspired by~\cite{meuleman2025fly, artdeco}, these primitives are initialized from \emph{keyframes} and \emph{mapper frames} by applying a Laplacian-of-Gaussian (LoG) probability map to prioritize high-frequency details and poorly reconstructed areas, while strictly excluding dynamic pixels via a motion map $\mathbf{M}$.
To mitigate overfitting, 80\% of the training views are sampled from historical frames before processing the next incoming frames. Upon sequence completion, a global optimization jointly refines all camera poses and Gaussian parameters to ensure trajectory consistency. More details are provided in the supplementary material. 

\section{Experiments}
\label{sec:exp}

\subsection{Experimental Setup}
\subsubsection{Datasets.}
We evaluate our approach on a diverse set of public benchmarks covering both indoor and outdoor environments. 
The indoor evaluation includes 38 scenes, consisting of 20 from ScanNet++~\cite{yeshwanth2023scannet++}, 10 from ScanNetV2~\cite{dai2017scannet} and 8 from VR-NeRF~\cite{xu2023vr}.
For outdoor environments, we use 22 large-scale sequences, including 8 from KITTI~\cite{geiger2012we}, 9 from Waymo~\cite{sun2020scalability}, and 5 from FAST-LIVO2~\cite{zheng2024fast}.

\begin{table}[t!]
\centering
\renewcommand{\arraystretch}{1.1} 
\setlength{\tabcolsep}{3.5pt}
\caption{\textbf{Tracking comparisons} in terms of ATE RMSE (m). 
We compare against state-of-the-art SLAM frameworks on both indoor and outdoor datasets. 
The best and second-best results are highlighted in \textbf{bold} and \underline{underline}, respectively.}
\label{tab:tracking_ate_comparison}

\resizebox{\linewidth}{!}{
\begin{tabular}{l|cccccc|c}
\toprule
Dataset & DROID-SLAM & MASt3R-SLAM & VGGT-SLAM &	VGGT-SLAM 2.0 & ARTDECO & Ours \ & \ Pi3X  \\
\midrule
ScanNet++ & 0.623 & 0.346 & 1.081 & 0.182 & \underline{0.137} & \textbf{0.065} \ & \ - \\
ScanNetV2       & 0.217 & 0.106 & 0.852 & \underline{0.073} & 0.141 & \textbf{0.051} \ & \ - \\
Waymo     & 7.535 & 16.610 & 13.166 & \underline{1.295} & 2.410 & \textbf{0.773} \ & \  0.827\\
KITTI     & 8.212 & 18.073 & 10.985 & \underline{2.521} & 2.938& \textbf{0.890} \ & \ 1.441 \\
\bottomrule
\end{tabular}}
\end{table}

\begin{table}[t]
\centering
\renewcommand{\arraystretch}{1.1}
\setlength{\tabcolsep}{2pt}
\caption{\textbf{Rendering comparisons} across indoor and outdoor datasets. 
We report visual quality metrics, the number of Gaussians, and average training time. 
The best and second-best results are highlighted in \textbf{bold} and \underline{underline}, respectively.}

\resizebox{\linewidth}{!}{
\begin{tabular}{l|ccc|ccc|ccc|cc}
\toprule
Indoor-dataset & \multicolumn{3}{c|}{ScanNet++} & \multicolumn{3}{c|}{ScanNetV2} & \multicolumn{3}{c|}{VR-NeRF} & \multicolumn{2}{c}{Efficiency} \\
Method & PSNR$\uparrow$ & SSIM$\uparrow$ & LPIPS$\downarrow$ 
       & PSNR$\uparrow$ & SSIM$\uparrow$ & LPIPS$\downarrow$ 
       & PSNR$\uparrow$ & SSIM$\uparrow$ & LPIPS$\downarrow$
       & \#GS & Time \\
\midrule
MonoGS        & 18.56 & 0.732 & 0.582 & 21.88 & 0.827 & 0.508 & 14.69 & 0.565 & 0.728 & 25.2k & 10.2 min \\
S3PO-GS       & 22.60 & 0.812 & 0.457 & 24.05 & 0.850 & 0.439 & 22.10 & 0.766 & 0.528 & 29.9k & 17.1 min \\
HI-SLAM2       & 23.89 & 0.811 & 0.342 & 23.75 & 0.820 & 0.295 & 28.25 & 0.868 & 0.277 & 152.1k & 6.9 min \\
\midrule
OnTheFly-NVS  & 22.00 & 0.808 & 0.336 & 23.38 & 0.861 & 0.348 & 28.23 & \underline{0.888} & 0.257 & 210.5k & 1.4 min \\
ARTDECO   & \underline{26.71} & \underline{0.864} & \underline{0.246} & \underline{26.03} & \underline{0.902} & \underline{0.278} & \underline{29.15} & 0.883 & \underline{0.231} & 936.7k & 5.1 min \\
\midrule
Ours    & \textbf{28.82} & \textbf{0.892} & \textbf{0.190} & \textbf{27.08} & \textbf{0.904} & \textbf{0.272} & \textbf{29.64} & \textbf{0.891} & \textbf{0.223} & 632.2k & 4.7 min \\
\bottomrule
\end{tabular}}

\resizebox{\linewidth}{!}{
\begin{tabular}{l|ccc|ccc|ccc|cc}
\toprule
Outdoor-dataset & \multicolumn{3}{c|}{KITTI} & \multicolumn{3}{c|}{Waymo} & \multicolumn{3}{c|}{FAST-LIVO2} & \multicolumn{2}{c}{Efficiency} \\
Method & PSNR$\uparrow$ & SSIM$\uparrow$ & LPIPS$\downarrow$ 
       & PSNR$\uparrow$ & SSIM$\uparrow$ & LPIPS$\downarrow$ 
       & PSNR$\uparrow$ & SSIM$\uparrow$ & LPIPS$\downarrow$
       & \#GS & Time \\
\midrule
MonoGS        & 14.60 & 0.490 & 0.761 & 19.49 & 0.742 & 0.636 & 18.37 & 0.589 & 0.719 & 27.2k & 13.5 min \\
S3PO-GS       & 18.45 & 0.609 & 0.416 & 25.09 & 0.809 & 0.437 & 20.62 & 0.645 & 0.499 & 109.7k & 24.4 min \\
HI-SLAM2       & 20.97 & 0.690 & 0.330 & 23.61 & 0.765 & 0.287 & 22.40 & 0.728 & 0.354 & 190.7k & 7.7 min \\
\midrule
OnTheFly-NVS  & 17.09 & 0.583 & 0.429 & 26.86 & 0.853 & 0.300 & 19.22 & 0.621 & 0.470 & 214.3k & 1.6 min \\
ARTDECO     & \underline{21.47} & \underline{0.709} & \underline{0.301} & \underline{28.63} & \underline{0.865} & \textbf{0.252} & \underline{24.71} & \underline{0.779} & \underline{0.294} & 1388.1k & 7.8 min \\
\midrule
Ours          & \textbf{22.47} & \textbf{0.727} & \textbf{0.288} & \textbf{28.94} & \textbf{0.880} & \underline{0.277} & \textbf{25.48} & \textbf{0.799} & \textbf{0.293} & 1098.9k & 7.2 min\\
\bottomrule
\end{tabular}}
\label{tab:quantitative_results}
\end{table}

\begin{figure}[t]
\includegraphics[width=\linewidth]{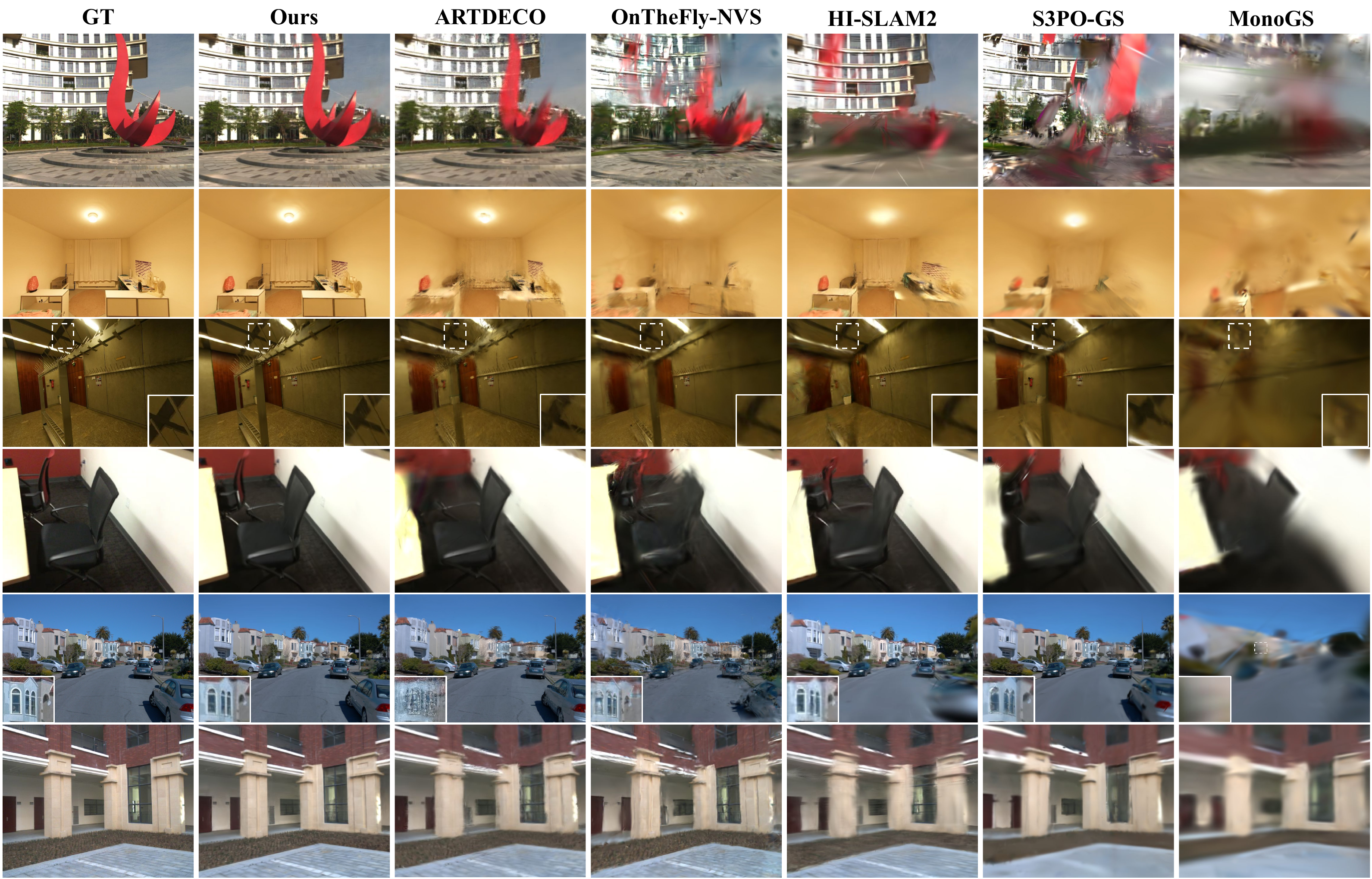}
\caption{\textbf{Qualitative comparisons} of rendering against on-the-fly reconstruction baselines across diverse datasets. M$^3$ preserves high-fidelity rendering details in challenging environments, particularly in regions highlighted by white rectangles.
}
\label{fig:main_results}
\end{figure}

\subsubsection{Metrics.}
We evaluate pose estimation accuracy, rendering quality, and efficiency. 
Pose accuracy is measured using the Absolute Trajectory Error (ATE) RMSE. 
Rendering quality is evaluated with PSNR, SSIM~\cite{wang2004image}, and LPIPS~\cite{zhang2018unreasonable}. 
Efficiency is assessed using the number of Gaussians and the training time.

\subsubsection{Baselines.}
We compare our approach with two categories of methods. For pose estimation, we evaluate against several SLAM frameworks, including DROID-SLAM~\cite{teed2021droid}, MASt3R-SLAM~\cite{murai2024_mast3rslam}, VGGT-SLAM~\cite{maggio2025vggt}, VGGT-SLAM 2.0~\cite{maggio2026vggt}, and ARTDECO~\cite{artdeco}. 
We also include the base model Pi3X~\cite{wang2025pi} without our enhancements as an additional baseline to evaluate the impact of proposed modifications.
For scene reconstruction, we compare against Gaussian Splatting-based streaming methods, including MonoGS~\cite{matsuki2024gaussian}, S3PO-GS~\cite{cheng2025outdoor}, HI-SLAM2~\cite{zhang2025hi}, OnTheFly-NVS~\cite{meuleman2025fly}, and ARTDECO~\cite{artdeco}. 
We further compare with feed-forward Gaussian Splatting approaches, including AnySplat~\cite{jiang2025anysplat} and Depth-Anything-3~\cite{lin2025depth}.


\subsection{Implementation Details.} We fine-tune the matching head of Pi3X for 200K iterations using the AdamW optimizer with a cosine learning rate schedule and a peak learning rate of $1\times10^{-4}$. Each training batch contains $N=8$ frames, and the balancing weight $\alpha$ in the training objective is set to $10.0$. In the M$^3$ framework, we use a sliding window of $L=8$ frames with up to four historical keyframes. The descriptor matching search radius is set to $r=4$, and the keyframe insertion threshold is defined as $\tau_k = \max(0.333W, 30)$. Following ARTDECO~\cite{artdeco} and OnTheFly-NVS~\cite{meuleman2025fly}, a 10k-iteration global refinement is performed after the streaming stage, while per-scene baselines are trained for 30k iterations. All experiments are conducted on an NVIDIA RTX 4090 GPU, while feed-forward baselines are evaluated on an NVIDIA H200 GPU due to memory requirements. For novel view synthesis evaluation, every eighth frame is held out and excluded from the mapper while its pose is optimized for evaluation. Additional training details are provided in the supplementary material.

\begin{table}[t]
\centering
\renewcommand{\arraystretch}{1.1}
\setlength{\tabcolsep}{2pt}
\caption{\textbf{Comparisons against feed-forward Gaussian Splatting methods.} 
We evaluate reconstruction quality and efficiency on fixed-length sequences. 
Our optimization-based approach consistently outperforms SOTA feed-forward models.}
\label{tab:feed_forward_horizontal}
\resizebox{\linewidth}{!}{
\begin{tabular}{l|cccc|cccc|cccc}
\toprule
Method & \multicolumn{4}{c|}{Ours} & \multicolumn{4}{c|}{Depth-Anything-3} & \multicolumn{4}{c}{AnySplat} \\
Dataset & PSNR$\uparrow$ & SSIM$\uparrow$ & LPIPS$\downarrow$ & GPU$\downarrow$ 
            & PSNR$\uparrow$ & SSIM$\uparrow$ & LPIPS$\downarrow$ & GPU $\downarrow$ 
            & PSNR$\uparrow$ & SSIM$\uparrow$ & LPIPS$\downarrow$ & GPU$\downarrow$ \\
\midrule
ScanNet++ & \textbf{27.789} & \textbf{0.873} & \textbf{0.211} & \textbf{22.4G} & 17.973 &  0.666 & 0.441 & 105.5G & \underline{22.337} & \underline{0.757} & \underline{0.236} & \underline{94.6G} \\
Waymo     & \textbf{28.346} & \textbf{0.870} & \textbf{0.273} & \textbf{23.2G} & 21.760 & 0.701 & \underline{0.338} & 82.0G & \underline{21.974} & \underline{0.729} & 0.341 & \underline{73.1G} \\
\bottomrule
\end{tabular}}
\end{table}

\begin{figure}[t]
\includegraphics[width=\linewidth]{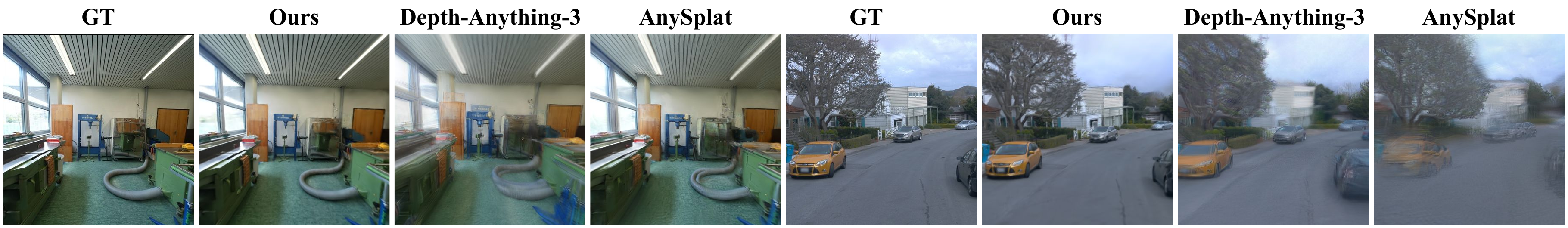}
\caption{\textbf{Qualitative comparisons of rendering against feed-forward Gaussian Splatting methods.} M$^3$ consistently preserves fine-grained visual details.
}
\label{fig:qualitative_ff}
\end{figure}

\subsection{Comparison}
\subsubsection{Pose Estimation Results Analysis}
Tab.~\ref{tab:tracking_ate_comparison} reports the pose estimation results on four indoor and outdoor benchmarks, comparing our method with recent SLAM frameworks and the base foundation model Pi3X~\cite{wang2025pi}. Our method achieves the lowest ATE on most evaluated sequences. 
This improvement can be attributed to the combination of strong geometric priors from the foundation model and geometric optimization within the SLAM framework, which together enable more accurate camera trajectory estimation. Additional qualitative trajectory visualizations are provided in the supplementary material.

\subsubsection{Reconstruction Results Analysis.}
In this section, we compare M$^3$ with two groups of reconstruction baselines: (i) SLAM-based Gaussian Splatting methods and (ii) feed-forward Gaussian Splatting methods. 
Tab.~\ref{tab:quantitative_results} reports results on six indoor and outdoor benchmarks for the SLAM-based setting. M$^3$ achieves consistently strong novel view synthesis quality while maintaining competitive efficiency. In particular, our method attains favorable rendering quality with a compact Gaussian representation 
and comparable training time. We attribute this improvement to improved pose estimation, which provides better-aligned camera trajectories and facilitates more stable and efficient 3DGS initialization. Fig.~\ref{fig:main_results} further shows qualitative comparisons, where M$^3$ produces sharper renderings with fewer artifacts and better preserves structural details. 

We further compare with feed-forward Gaussian Splatting methods. 
Tab.~\ref{tab:feed_forward_horizontal} and Fig.~\ref{fig:qualitative_ff} report results under the AnySplat protocol~\cite{jiang2025anysplat}: we randomly sample $256$ images from ScanNet++ and $199$ images from Waymo and apply the standard center-cropping procedure. 
M$^3$ achieves better reconstruction quality than feed-forward baselines on both indoor and outdoor sets, and remains stable in large-scale scenes. Moreover, our method requires less memory than feed-forward baselines, facilitating deployment on commercial hardware. Additional results are provided in the supplementary material.

\begin{figure}[t]
\includegraphics[width=\linewidth]{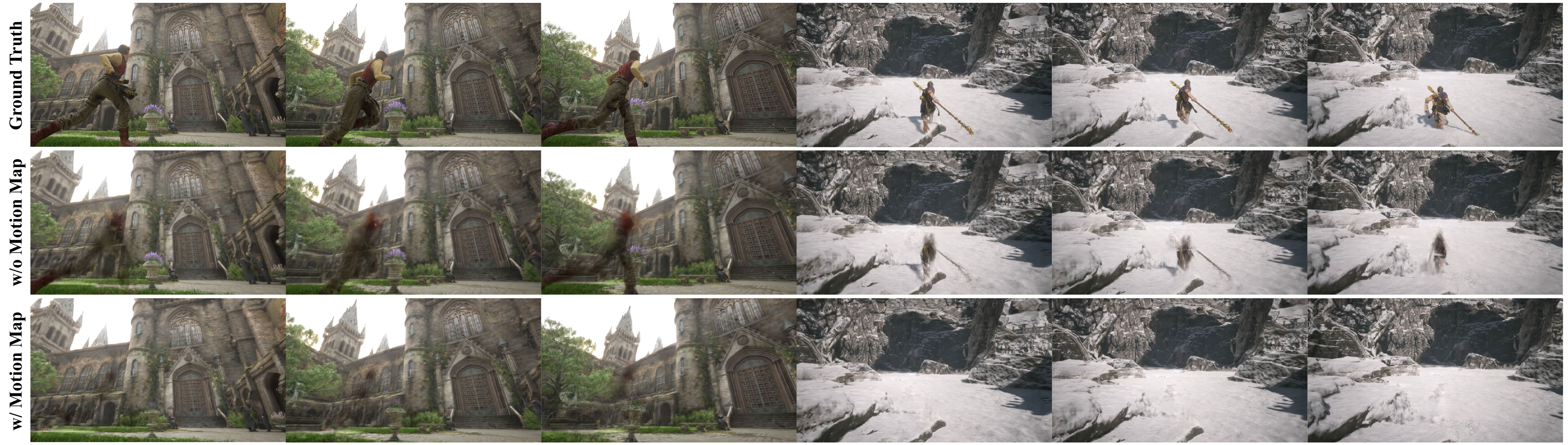}
\caption{\textbf{Effect of the Motion Map in dynamic environments.} 
By detecting dynamic regions, the Motion Map enables moving objects to be excluded from the static reconstruction, thereby improving structural consistency in dynamic scenes.
}
\label{fig:dynamic}
\end{figure}

\subsection{Dynamic Scene Reconstruction}
Fig.~\ref{fig:dynamic} illustrates the effect of the proposed Motion Map in dynamic environments. Without the Motion Map, dynamic objects are often reconstructed as blurry ghosting artifacts, which degrades both rendering quality and geometric accuracy. 
By identifying and suppressing dynamic regions, the Motion Map allows moving objects (e.g., pedestrians) to be excluded from the static reconstruction process. 
As a result, the reconstructed scene remains temporally consistent and less affected by transient motion.


\begin{table}[t!]
\centering
\renewcommand{\arraystretch}{1.1} 
\setlength{\tabcolsep}{3pt}
\caption{\textbf{Ablation studies on the ScanNet++ dataset} evaluating the effect of key design choices and hyperparameters.}
\label{tab:ablation}

\resizebox{\linewidth}{!}{
\begin{tabular}{l|ccccccc}
\toprule
Metrics & Full  & w/o descriptor & w/o intr. align & w/o global opt. & $\rm{pi3x} \rightarrow \rm{pi3}$ & split front\&back-end & $L$=16  \\
\midrule
ATE$\downarrow$  & \textbf{0.065} & 0.094 & 0.236 & 0.130 & 0.068 & 0.098 & 0.077 \\
PSNR$\uparrow$ & \textbf{28.82} & 27.73 & 26.14 & 28.49 & 28.77 & 28.41 & 28.57 \\
SSIM$\uparrow$ & \textbf{0.892} & 0.874 & 0.851 & 0.886 & \textbf{0.892} & 0.886 & 0.888 \\
LPIPS$\downarrow$ & \textbf{0.190} & 0.221 & 0.258 & 0.201 & 0.191 & 0.203 & 0.199 \\
\bottomrule
\end{tabular}}
\end{table}

\subsection{Ablation Study}
Tab.~\ref{tab:ablation} presents ablation results on the ScanNet++ dataset evaluating the effect of key design choices and hyperparameters. (i) Setting the descriptor search radius to $r=0$ degrades pose estimation and rendering quality due to the loss of fine-grained correspondences. (ii) Removing per-inference intrinsic alignment weakens coarse matching, leading to less consistent geometry between predictions and the global map. (iii) Disabling global optimization reduces localization accuracy, highlighting its role in mitigating drift. (iv) Replacing the Pi3X backbone with Pi3 results in a small performance drop across metrics, indicating the benefit of stronger Pi3X priors for reconstruction. (v) Decoupling frontend and backend introduces redundant inference passes, reducing stability and increasing training time compared with our integrated streaming framework. (vi) Finally, varying the sliding window size shows that $L=8$ provides a good balance between temporal stability and computational cost. More details are provided in the supplementary material.


\section{Limitations}
\label{sec:limitation}
Although M$^3$ tightly integrates foundation-model priors into a SLAM framework, the framework still relies on the correctness of feed-forward predictions. When the foundation model produces severely inaccurate correspondences or geometric priors, the SLAM optimization may fail to recover, as the current framework lacks a dedicated fallback mechanism. Moreover, the framework currently operates purely in a monocular visual setting and does not leverage complementary sensing modalities such as LiDAR or inertial measurements. Multi-sensor fusion could further improve robustness and accuracy in challenging scenarios.
\section{Conclusion}
\label{sec:conclusion}
We presented M$^3$, an efficient and robust streaming reconstruction framework for uncalibrated monocular video. The key insight is to enhance a multi-view foundation model, Pi3X~\cite{wang2025pi}, with pixel-level dense matching and tightly incorporate it into a SLAM pipeline. This design enables consistent dense correspondences for pose optimization, reduces redundant model inferences, and supports stable tracking together with high-fidelity 3DGS reconstruction in long video streams. To better handle real-world complexities, M$^3$ further incorporates descriptor-based dynamic region suppression and intrinsic alignment to mitigate drift and improve global consistency. Extensive experiments on diverse indoor and outdoor benchmarks demonstrate that M$^3$ achieves state-of-the-art pose estimation and reconstruction quality with competitive efficiency. We believe M$^3$ represents a promising step toward practical and scalable streaming reconstruction.

\bibliographystyle{splncs04}
\bibliography{main}

\section{Supplementary Material}

This supplementary material provides additional technical details and experimental results to complement the main paper. We first delineate the architecture of our enhanced Pi3X model in Sec.~\ref{sec:model_arch}. Sec.~\ref{sec:sliding_window} then elaborates on the loop closure mechanism within the sliding window. In Sec.~\ref{sec:neural_gaussians}, we describe the initialization and training strategies for the neural Gaussian representation. Sec.~\ref{sec:ablation} presents an ablation study on the impact of various multi-view foundation models applied to our framework, followed by comprehensive implementation details in Sec.~\ref{sec:implementation}, including the fine-tuning process and SLAM integration. Finally, Sec.~\ref{sec:exp} showcases extended qualitative results along with detailed per-scene quantitative evaluations. Furthermore, we provide a supplementary demo to demonstrate our pose estimation and continuous rendering performance in real-world scenes, highlighting the ability to perform rapid reconstruction from uncalibrated video.

\subsection{Details of Multi-view Foundation Model}
\label{sec:model_arch}

The multi-view foundation model we build upon is Pi3X~\cite{wang2025pi}, a permutation-equivariant feed-forward network designed for multi-view geometry reconstruction. The model architecture follows a modular design consisting of a feature encoder, a multi-view interaction module, and multiple task-specific prediction heads. This structure allows the framework to jointly process an uncalibrated set of images and infer consistent geometric priors without anchoring to a fixed reference view.

Specifically, for a batch size $B$, the input consists of $N$ monocular images with a resolution of $H \times W$, forming an input tensor $\mathcal{I} \in \mathbb{R}^{B \times N \times 3 \times H \times W}$. These images are first processed by a shared feature encoder, typically a DINOv2 backbone with a patch size of $14 \times 14$, to extract $L = (H/14) \times (W/14)$ tokens per image with an embedding dimension $D=1024$, resulting in intermediate features of dimension $\mathbb{R}^{B \times N \times L \times D}$. These tokens are then passed into the multi-view interaction module, which employs several transformer layers with cross-view attention to aggregate geometric information across all $N$ views while maintaining permutation equivariance. The refined tokens are subsequently fed into specialized heads for geometric and camera parameter estimation. The pose head utilizes global pooling and linear layers to predict camera rotation $R \in \mathbb{R}^{B \times N \times 9}$ and translation $t \in \mathbb{R}^{B \times N \times 3}$ for each view. Parallel to this, the confidence head produces an uncertainty map with dimensions $\mathbb{R}^{B \times N \times H \times W \times 1}$, while the metric head outputs a global scalar $\mathcal{S} \in \mathbb{R}^{B}$ to provide metric scale information for the batch. To enable the fine-grained geometric constraints required for the M$^3$ pipeline, we introduce an additional matching head that outputs pixel-level descriptors in $\mathbb{R}^{B \times N \times H \times W \times 24}$ along with a corresponding matching confidence map in $\mathbb{R}^{B \times N \times H \times W \times 1}$.

\subsection{Details of Loop Closure}
\label{sec:sliding_window}

During the image retrieval, a loop closure is triggered when the retrieved keyframes exhibit significant temporal discontinuities relative to the current sequence. Following ARTDECO~\cite{artdeco}, we retain the top $N_c$ keyframe candidates identified by the SALAD~\cite{izquierdo2024optimal}. These candidates, along with the current reference frame, are then processed through the multi-view foundation model, Pi3X~\cite{wang2025pi}, to evaluate their geometric consistency. To ensure the reliability of the loop constraints, we rank these candidates based on their matching ratios and select the top $K-1$ keyframes as the final retrieved set. Subsequently, the factor graph is updated by establishing edges between the reference frame and these newly identified keyframes. Finally, a global Bundle Adjustment (BA) is performed to optimize the entire trajectory and scene representation, effectively eliminating accumulated drift and ensuring global consistency.

\subsection{Details of Neural Gaussian}
\label{sec:neural_gaussians}

\subsubsection{Gaussian Initialization.} Our initialization strategy focuses on spatially adaptive density control to ensure high-fidelity reconstruction without incurring prohibitive memory overhead. Unlike traditional 3DGS methods~\cite{kerbl20233d} that utilize isotropic point clouds, we employ a probabilistic selection mechanism to identify regions requiring geometric refinement inspired by~\cite{lu2024scaffold, ren2024octree}. When a mapper frame or keyframe is processed by the backend, we compute an insertion probability $P_a(u, v)$ for each pixel $(u, v)$ based on the structural discrepancy between the ground-truth image $I$ and the current rendered view $\tilde{I}$. Following~\cite{meuleman2025fly}, we utilize the Laplacian of Gaussian (LoG) operator to prioritize high-frequency textures and poorly reconstructed geometries:
\begin{equation}
P_a(u, v) = \max \left( \min(| \nabla^2(G_\sigma) * I(u,v) |, 1) - \min(| \nabla^2(G_\sigma) * \tilde{I}(u,v) |, 1), 0 \right),
\end{equation}
where $G_\sigma$ denotes a Gaussian smoothing kernel. A new Gaussian primitive is instantiated only if $P_a(u, v)$ exceeds a predefined activation threshold $\tau_a$.
Once a candidate pixel is selected, we initialize its corresponding 3D Gaussian primitive defined by its center $\mu$, spherical harmonics (SH) coefficients, opacity $\alpha$, scale $s$ and rotation $r$. The center $\mu$ and zero-order SH coefficients are directly anchored to the pointmap coordinates and pixel colors retrieved from the foundation model. To suppress noise in uncertain regions, the initial opacity is weighted by the backend confidence score $C(u, v)$ as $\alpha = 0.2 \cdot C(u, v)$. The base scale $s$ is derived from the local pixel intensity and camera geometry as follows:
\begin{equation}
s = \frac{d_i \cdot s'}{f}, \quad s' = \frac{1}{2 \sqrt{\min(| (\nabla^2 G_\sigma) * I(u, v) |, 1)}},\end{equation}
where $d_i$ represents the distance from the Gaussian center to the optical center and $f$ is the focal length. Here, $s'$ acts as an image-space scale factor representing the local point density. To further refine the anisotropy of the primitives, we employ two lightweight MLPs to predict residual scale and rotation:
\begin{equation}
s_f = s \cdot \text{MLP}_s(f_g \oplus f_l), \quad r_f = r \cdot \text{MLP}_r(f_g \oplus f_l),
\end{equation}
where $f_l$ is an individual feature and $f_g$ is a region feature capturing the local voxel context. This hybrid parameterization ensures that each Gaussian maintains local distinctiveness while remaining globally consistent with its neighbors.

\subsubsection{Level of Detail Design.} To accommodate large-scale environments, we organize the Gaussian field into a hierarchical Level of Detail (LoD) structure~\cite{artdeco}. Each Gaussian is assigned a level $l \in \{0, \dots, L-1\}$, where level 0 represents the highest resolution. Primitives at level $l$ correspond to a pixel patch of size $2^{2l}$ in the original image. During initialization, we assign each Gaussian a distance-dependent visibility parameter $d_{max} = d \cdot 2^{2l}$, where $d$ is the depth at the time of creation. During the rendering process, the opacity is smoothly modulated to prevent flickering artifacts: a Gaussian is fully active if the current distance $d_r \leq d_{max}$, and its opacity $\alpha$ is linearly faded to zero as $d_r$ transitions from $d_{max}$ to $2d_{max}$. This distance-aware management enables stable rendering across varying viewing distances while preserving computational efficiency in navigable spaces.

\subsubsection{Training Strategy.} We adopt a staged optimization scheme to balance real-time tracking requirements with reconstruction quality. For streaming input, whenever a mapper frame or keyframe is introduced, we perform $K=20$ optimization iterations to incorporate new Gaussians and refine existing geometry. In contrast, common frames, which do not trigger new Gaussian insertion, undergo only $K/2$ iterations to update appearance and refine poses. To prevent local overfitting during the streaming phase, training frames are sampled stochastically, with a 20\% probability of selecting the current frame and an 80\% probability of selecting historical frames.
Upon completion of the video sequence, we execute a global refinement phase. During this stage, sampling probabilities are biased toward frames that received fewer updates during the streaming process. Consistent with standard SLAM practices~\cite{matsuki2024gaussian}, camera poses are jointly optimized alongside Gaussians. Gradients from the photometric loss are propagated back to the camera translation and rotation, ensuring that the final trajectory and the scene representation are perfectly aligned in a globally consistent metric space.

\subsection{Ablations on Foundation Models}
\label{sec:ablation}

While the ablation study in the main text highlights that Pi3X~\cite{wang2025pi} delivers superior pose estimation performance compared to its predecessor Pi3, we further investigate the generalizability and effectiveness of other multi-view foundation models within the M$^3$ framework. Specifically, we evaluate the integration of Depth-Anything-3 (DA3)~\cite{lin2025depth} and VGGT~\cite{wang2025vggt} as alternative backbones. To ensure a fair comparison, we apply a consistent fine-tuning strategy to these models, enabling their respective matching heads to generate descriptors and confidence maps of the same dimensionality for dense matching.

Quantitative results on the ScanNet++ dataset are summarized in Table~\ref{tab:foundation_models}. The experimental data indicate that Pi3X and DA3 emerge as the leading backbones, significantly outperforming the earlier VGGT model. However, while DA3 provides strong individual geometric priors, its consistency across multiple inference passes is slightly inferior to that of Pi3X. When integrated into our unified SLAM optimization pipeline, this relative lack of temporal-geometric consistency in DA3 leads to marginally higher pose drift (ATE) and reduced reconstruction fidelity compared to the Pi3X-based implementation. These comparisons underscore the importance of selecting a foundation model with robust cross-view interaction and consistent multi-view alignment to facilitate high-precision monocular tracking and mapping.

\begin{table}[t]
\centering
\renewcommand{\arraystretch}{1.1}
\setlength{\tabcolsep}{2pt}
\caption{\textbf{Comparison of Multi-view Foundation Models within the M$^3$ Framework.} 
We evaluate the performance of different backbones on the ScanNet++ dataset~\cite{yeshwanth2023scannet++}. 
Pi3X~\cite{wang2025pi} and Depth-Anything-3~\cite{lin2025depth} represent the state-of-the-art in geometric priors, 
with Pi3X achieving the highest accuracy due to its superior cross-inference consistency.}
\vspace{-6pt}
\label{tab:foundation_models}
\resizebox{\linewidth}{!}{
\begin{tabular}{l|cccc|cccc|cccc}
\toprule
Method & \multicolumn{4}{c|}{Ours (Pi3X)} & \multicolumn{4}{c|}{Depth-Anything-3} & \multicolumn{4}{c}{VGGT} \\
Dataset & ATE$\downarrow$ & PSNR$\uparrow$ & SSIM$\uparrow$ & LPIPS$\downarrow$  
            & ATE$\downarrow$ & PSNR$\uparrow$ & SSIM$\uparrow$ & LPIPS$\downarrow$  
            & ATE$\downarrow$ & PSNR$\uparrow$ & SSIM$\uparrow$ & LPIPS$\downarrow$  \\
\midrule
ScanNet++ & \textbf{0.065} & \textbf{28.82} & \textbf{0.892} & \textbf{0.190} & \underline{0.079} & \underline{28.09} & \underline{0.882} & \underline{0.211} & 0.106 & 27.33 & 0.869 & 0.234 \\
\bottomrule
\end{tabular}}
\vspace{-6pt}
\end{table}

\subsection{More Implementation Details}
\label{sec:implementation}
\subsubsection{Details of Pi3X Finetuning.}
The matching head is trained for 200K iterations using the AdamW optimizer across the diverse datasets summarized in Table 2. We employ a cosine learning rate scheduler with a peak learning rate of $1 \times 10^{-4}$ and a linear warm-up phase of 1K steps.
During training, each input batch consists of $N=8$ frames. To enhance the model's multi-scale robustness, we apply random resolution scaling within a ratio of $[0.8, 1.2]$, while maintaining the long edge of the images at 518 pixels. For each source frame, we randomly sample 8,192 query points to compute the matching loss. The balancing hyper-parameter $\alpha$ in the total training objective is set to 10.0. The entire training process was conducted on 16 NVIDIA RTX 4090 GPUs, taking approximately 4 days to reach convergence. More training details are provided in the supplementary material.

\subsubsection{Details of ARTDECO-V2 Framework.}
For the sliding window, we employ $L=8$ frames, including $K = \min(N_k, 4)$ historical keyframes, where $N_k$ denotes the total keyframe count. We set the search radius for descriptor-based matching to $r=4$ and define the keyframe insertion threshold as $\tau_k = \max(0.333 \cdot W, 30)$. Loop closure is performed across $N_c = \min(23, N_k)$ candidate keyframes to ensure global consistency.

\subsection{Additional Qualitative and Quantitative Results}
\label{sec:exp}

\begin{figure}[htbp]
\includegraphics[width=\linewidth]{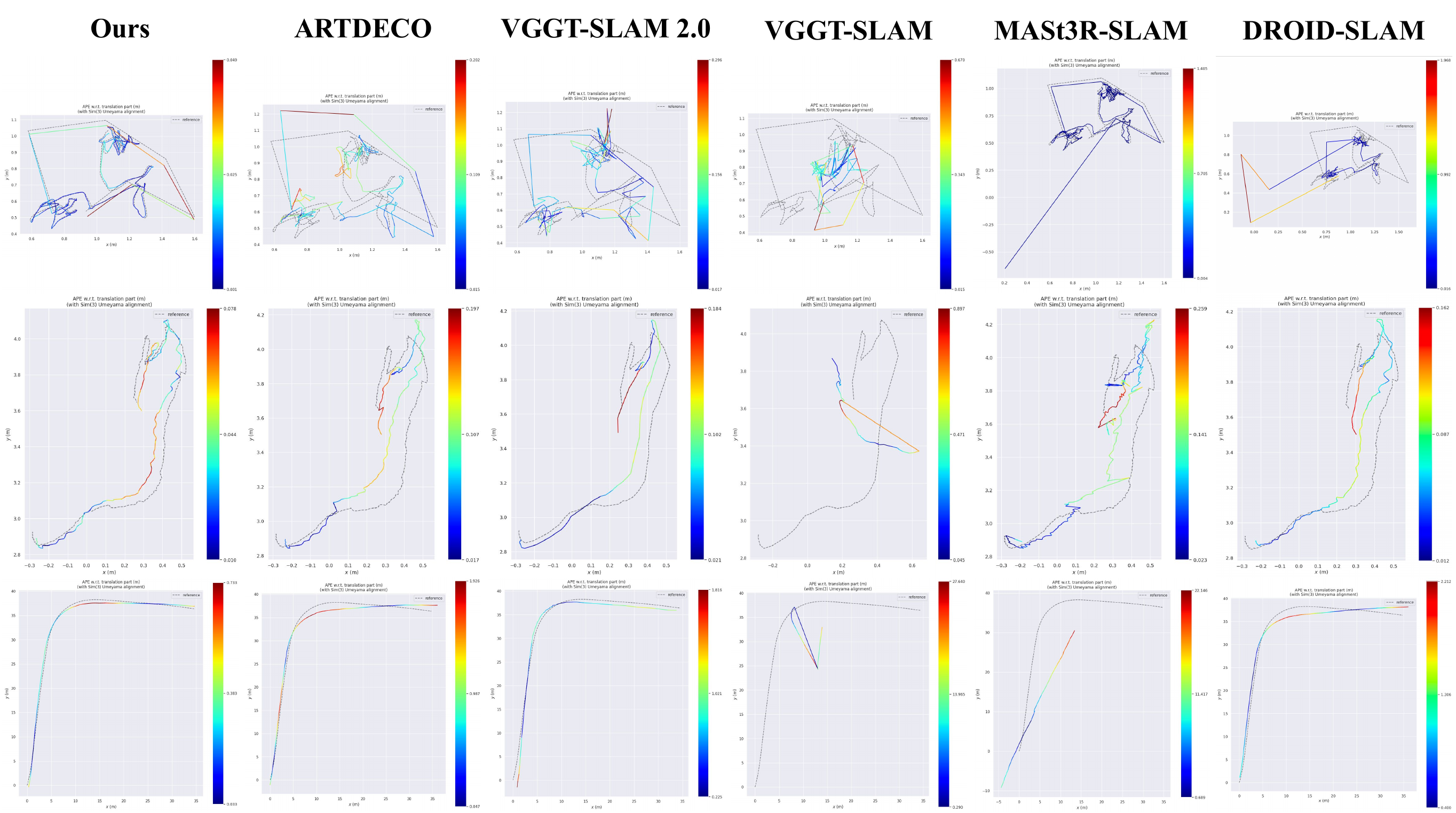}
\caption{\textbf{Qualitative comparisons} of camera trajectories against on-the-fly reconstruction baselines across diverse datasets. Our visualized trajectories align most closely with the ground truth (GT), exhibiting no significant divergence or inconsistency.
}
\vspace{-6pt}
\label{fig:suppl3}
\vspace{-12pt}
\end{figure}

\begin{figure}[htbp]
\includegraphics[width=\linewidth]{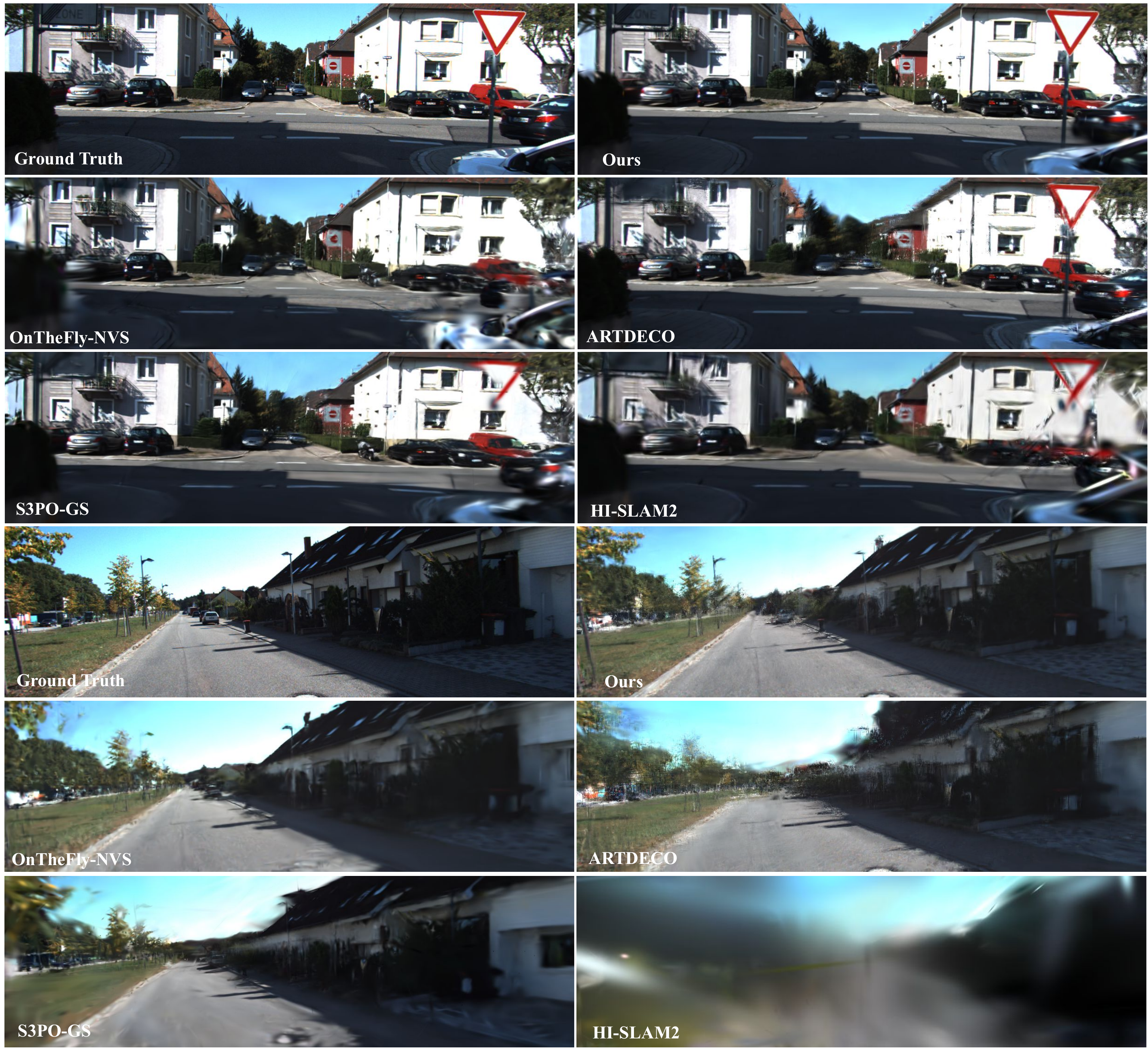}
\caption{\textbf{Qualitative comparisons} of rendering against on-the-fly reconstruction baselines in outdoor scenes. Our rendered outputs exhibit superior clarity and are devoid of visual artifacts.
}
\label{fig:suppl2}
\end{figure}

\begin{figure}[htbp]
\includegraphics[width=\linewidth]{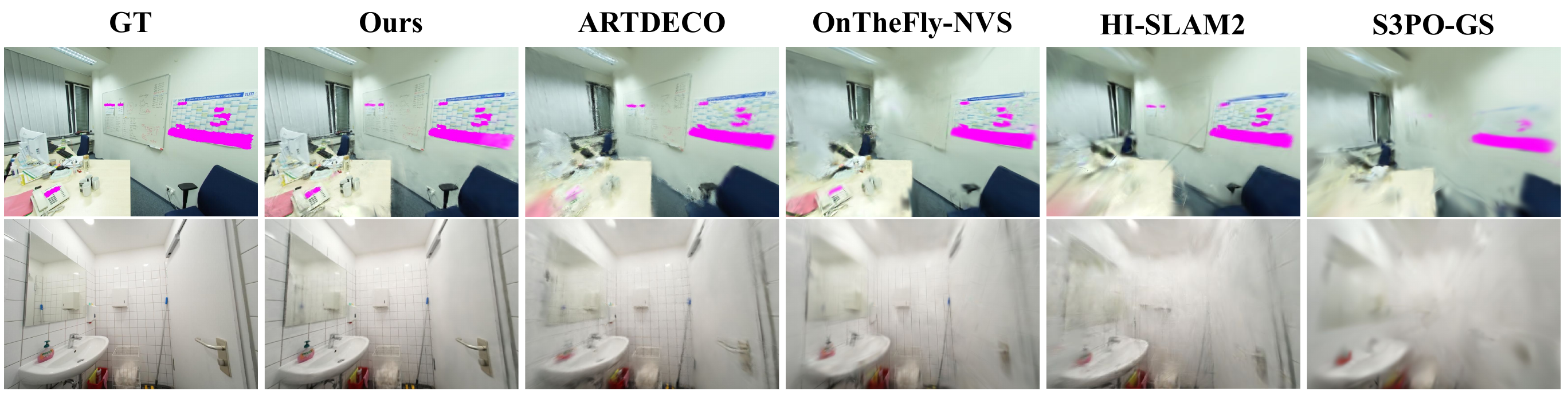}
\caption{\textbf{Qualitative comparisons} of rendering against on-the-fly reconstruction baselines in indoor scenes. Our rendered outputs exhibit superior clarity and are devoid of visual artifacts.
}
\label{fig:suppl1}
\end{figure}

In this section, we provide extended qualitative comparisons between M$^3$ and baseline methods~\cite{artdeco, meuleman2025fly, maggio2026vggt, maggio2025vggt, murai2024_mast3rslam, teed2021droid, wang2025pi, matsuki2024gaussian, cheng2025outdoor, zhang2025hi} in Figs.~\ref{fig:suppl3}, \ref{fig:suppl2}, and \ref{fig:suppl1}, showcasing rendered views and estimated camera trajectories across diverse indoor and outdoor environments. Furthermore, we report comprehensive per-scene quantitative evaluations to validate the performance of our framework in Tabs.~\ref{tab:track_scannetpp_rmse} to \ref{tab:vrnerf_lpips}. Specifically, trajectory accuracy is measured via Absolute Trajectory Error (ATE RMSE) on ScanNet++~\cite{yeshwanth2023scannet++}, ScanNetV2~\cite{dai2017scannet}, Waymo~\cite{sun2020scalability}, and KITTI~\cite{geiger2012we} datasets. We also evaluate scene reconstruction fidelity using standard image synthesis metrics (PSNR, SSIM~\cite{wang2004image}, and LPIPS~\cite{zhang2018unreasonable}) across ScanNet++~\cite{yeshwanth2023scannet++}, ScanNetV2~\cite{dai2017scannet}, VR-NeRF~\cite{xu2023vr}, Waymo~\cite{sun2020scalability}, KITTI~\cite{geiger2012we}, and Fast-LIVO2~\cite{zheng2024fast}. These detailed results further demonstrate the robustness and precision of M$^3$ in handling complex real-world sequences.

\begin{table}[htbp]
\begin{center}
\renewcommand{\arraystretch}{1.1}
\setlength{\tabcolsep}{2.5pt}
\footnotesize
\caption{ATE RMSE (m) on the ScanNet++ Dataset} 
\label{tab:track_scannetpp_rmse}%
\resizebox{1\linewidth}{!}{
\begin{tabular}{c|ccccccc}
\toprule
Method & 00777c41d4 & 126d03d821 & 1cbb105c6a & 4808c4a397 & 546292a9db & 7543973e1a & 0a7cc12c0e \\
\midrule
DROID-SLAM & 1.677 & 0.231 & 0.304 & 0.253 & 0.386 & 0.414 & 1.285 \\
MASt3R-SLAM & 0.584 & \underline{0.170} & 0.232 & 0.338 & 0.144 & \underline{0.132} & 0.687 \\
VGGT-SLAM & 1.715 & 1.765 & 0.651 & 1.594 & 0.904 & 1.185 & 1.688 \\
VGGT-SLAM 2.0 & 0.202 & 0.178 & 0.185 & 0.273 & \textbf{0.037} & 0.333 & \underline{0.207} \\
ARTDECO & \textbf{0.072} & \textbf{0.057} & \underline{0.148} & \underline{0.086} & \underline{0.043} & 0.176 & 0.455 \\
Ours & \underline{0.073} & 0.305 & \textbf{0.025} & \textbf{0.054} & 0.087 & \textbf{0.053} & \textbf{0.016} \\
\midrule
Method & 0d8ead0038 & 1a8e0d78c0 & 21532e059d & 251443268c & 3f15a9266d & 617dc40bca & 7079b59642 \\
\midrule
DROID-SLAM & 0.120 & 1.261 & 0.062 & 1.048 & 0.489 & 0.149 & 0.748 \\
MASt3R-SLAM & \textbf{0.031} & 0.505 & \textbf{0.019} & 0.505 & \underline{0.061} & \textbf{0.062} & 0.651 \\
VGGT-SLAM & 0.339 & 1.390 & 0.334 & 1.241 & 0.616 & 0.279 & 1.768 \\
VGGT-SLAM 2.0 & 0.133 & 0.178 & 0.076 & 0.175 & 0.158 & 0.131 & 0.194 \\
ARTDECO & \underline{0.046} & \underline{0.161} & 0.067 & \underline{0.094} & 0.097 & \underline{0.101} & \underline{0.191} \\
Ours & 0.104 & \textbf{0.027} & \underline{0.059} & \textbf{0.038} & \textbf{0.015} & 0.129 & \textbf{0.059} \\
\midrule
Method & 8b2c0938d6 & 8be0cd3817 & ad2d07fd11 & b73f5cdc41 & e814070fd4 & f1efd25854 &  \\
\midrule
DROID-SLAM & 0.901 & 1.327 & 0.676 & 0.237 & 0.706 & 0.177 &  \\
MASt3R-SLAM & 0.284 & 1.036 & 1.083 & 0.116 & 0.199 & \underline{0.072} &  \\
VGGT-SLAM & 1.366 & 1.521 & 1.070 & 1.000 & 0.824 & 0.362 &  \\
VGGT-SLAM 2.0 & \underline{0.265} & 0.177 & 0.258 & \underline{0.091} & 0.252 & 0.138 &  \\
ARTDECO & 0.326 & \underline{0.108} & \underline{0.125} & 0.140 & \underline{0.114} & 0.131 &  \\
Ours & \textbf{0.061} & \textbf{0.048} & \textbf{0.051} & \textbf{0.024} & \textbf{0.054} & \textbf{0.027} &  \\
\bottomrule
\end{tabular}%
}
\end{center}
\end{table}

\begin{table}[htbp]
\begin{center}
\renewcommand{\arraystretch}{1.1}
\setlength{\tabcolsep}{7pt}
\footnotesize
\caption{ATE RMSE (m) on the ScanNetV2 Dataset} 
\label{tab:scannetv2_rmse}%
\resizebox{1\linewidth}{!}{
\begin{tabular}{c|ccccc}
\toprule
Method & scene0081\_00 & scene0354\_00 & scene0658\_00 & scene0671\_00 & scene0441\_00 \\
\midrule
DROID-SLAM & \underline{0.084} & 0.529 & 0.171 & \underline{0.032} & 0.068 \\
MASt3R-SLAM & 0.130 & 0.150 & 0.094 & 0.084 & 0.107 \\
VGGT-SLAM & 0.447 & 1.131 & 0.553 & 1.191 & 0.422 \\
VGGT-SLAM 2.0 & 0.097 & \textbf{0.063} & 0.080 & 0.036 & \underline{0.047} \\
ARTDECO & 0.106 & 0.543 & \underline{0.060} & 0.039 & 0.052 \\
Ours & \textbf{0.046} & \underline{0.093} & \textbf{0.044} & \textbf{0.029} & \textbf{0.040} \\
\midrule
Method & scene0461\_00 & scene0423\_00 & scene0144\_00 & scene0559\_00 & scene0316\_00 \\
\midrule
DROID-SLAM & 0.046 & \underline{0.068} & 0.099 & 0.975 & 0.095 \\
MASt3R-SLAM & 0.073 & 0.085 & \underline{0.078} & 0.158 & 0.103 \\
VGGT-SLAM & 0.793 & 1.388 & 0.668 & 1.494 & 0.429 \\
VGGT-SLAM 2.0 & \underline{0.041} & 0.074 & 0.082 & 0.140 & \underline{0.070} \\
ARTDECO & 0.047 & 0.197 & 0.104 & \underline{0.138} & 0.124 \\
Ours & \textbf{0.032} & \textbf{0.046} & \textbf{0.061} & \textbf{0.069} & \textbf{0.046} \\
\bottomrule
\end{tabular}%
}
\end{center}
\end{table}

\begin{table}[htbp]
\begin{center}
\renewcommand{\arraystretch}{1.1}
\setlength{\tabcolsep}{5.5pt}
\footnotesize
\caption{ATE RMSE (m) on the Waymo Dataset} 
\label{tab:waymo_rmse}%
\resizebox{1\linewidth}{!}{
\begin{tabular}{c|ccccccccc}
\toprule
Method & 100613 & 106762 & 132384 & 13476 & 152706 & 153495 & 158686 & 163453 & 405841 \\
\midrule
DROID-SLAM & 1.490 & 2.117 & 3.178 & 1.163 & 2.560 & 12.146 & 11.144 & 24.172 & 9.848 \\
MASt3R-SLAM & 17.464 & 32.556 & 48.649 & 13.496 & 22.706 & 4.029 & 3.278 & 4.548 & 2.767 \\
VGGT-SLAM & 25.969 & \textbf{1.247} & \underline{1.607} & 12.910 & 33.078 & 17.520 & 14.157 & \underline{1.455} & 10.554 \\
VGGT-SLAM 2.0 & \underline{0.927} & \underline{1.516} & 1.756 & 1.710 & \textbf{0.620} & \underline{1.029} & \underline{1.677} & 1.543 & \underline{0.880} \\
ARTDECO & 1.828 & 3.682 & 3.175 & \underline{1.139} & 1.103 & 3.018 & 2.606 & 3.763 & 1.380 \\
Ours & \textbf{0.453} & 1.621 & \textbf{0.872} & \textbf{0.356} & \underline{0.639} & \textbf{0.974} & \textbf{0.204} & \textbf{1.369} & \textbf{0.471} \\
\hline
Pi3X & 0.730 & 0.938 & 2.343 & 0.509 & 0.415 & 0.933 & 0.361 & 0.719 & 0.492 \\
\bottomrule
\end{tabular}%
}
\end{center}
\end{table}

\begin{table}[htbp]
\begin{center}
\renewcommand{\arraystretch}{1.1}
\setlength{\tabcolsep}{8pt}
\footnotesize
\caption{ATE RMSE (m) on the KITTI Dataset} 
\label{tab:KITTI_rmse}%
\resizebox{1\linewidth}{!}{
\begin{tabular}{c|cccccccc}
\toprule
Method & 00 & 02 & 03 & 05 & 06 & 07 & 08 & 10 \\
\midrule
DROID-SLAM & 18.534 & 3.510 & 5.335 & 8.316 & 10.982 & 6.045 & 8.555 & 4.417 \\
MASt3R-SLAM & 26.038 & 56.199 & \underline{1.519} & 21.461 & 1.621 & 1.943 & 31.022 & 4.783 \\
VGGT-SLAM & 2.205 & 35.429 & 2.137 & 1.824 & 20.523 & 22.128 & 0.543 & 3.093 \\
VGGT-SLAM 2.0 & 2.300 & 3.751 & 2.213 & \underline{1.811} & \underline{3.374} & 3.212 & \underline{0.557} & 2.952 \\
ARTDECO & \underline{1.913} & \underline{3.144} & 1.790 & 2.552 & 8.136 & \underline{1.745} & 1.962 & \underline{2.259} \\
Ours & \textbf{0.376} & \textbf{1.284} & \textbf{0.944} & \textbf{1.062} & \textbf{1.395} & \textbf{0.511} & \textbf{0.465} & \textbf{1.083} \\
\hline
Pi3X & 1.239 & 5.906 & 0.539 & 0.802 & 0.620 & 0.781 & 0.532 & 1.105 \\
\bottomrule
\end{tabular}%
}
\end{center}
\end{table}

\begin{table}[htbp]
\begin{center}
\renewcommand{\arraystretch}{1.1}
\setlength{\tabcolsep}{2.5pt}
\footnotesize
\caption{PSNR on the ScanNet++ Dataset} 
\label{tab:track_scannetpp_psnr}%
\resizebox{1\linewidth}{!}{
\begin{tabular}{c|ccccccc}
\toprule
Method & 00777c41d4 & 126d03d821 & 1cbb105c6a & 4808c4a397 & 546292a9db & 7543973e1a & 0a7cc12c0e \\
\midrule
MonoGS & 11.57 & 21.14 & 21.64 & 21.98 & 11.20 & 21.37 & 18.51 \\
S3PO-GS & 17.78 & 23.38 & 25.23 & 22.64 & 20.13 & 24.21 & 20.67 \\
HI-SLAM2 & 18.45 & 25.51 & 23.09 & 23.95 & 22.77 & 26.79 & 23.35 \\
OnTheFly-NVS & 19.49 & 19.43 & 19.52 & 18.57 & \textbf{26.13} & 20.86 & 21.55 \\
ARTDECO & \textbf{25.33} & \underline{28.64} & \underline{26.93} & \textbf{30.17} & \underline{25.53} & \underline{27.69} & \underline{28.88} \\
Ours & \underline{25.20} & \textbf{29.56} & \textbf{28.29} & \underline{30.12} & 23.43 & \textbf{30.67} & \textbf{31.14} \\
\midrule
Method & 0d8ead0038 & 1a8e0d78c0 & 21532e059d & 251443268c & 3f15a9266d & 617dc40bca & 7079b59642 \\
\midrule
MonoGS & 24.95 & 9.59 & 22.44 & 17.58 & 22.42 & 24.23 & 21.20 \\
S3PO-GS & 25.99 & 18.29 & 21.06 & 21.06 & 22.80 & 25.26 & 24.98 \\
HI-SLAM2 & \underline{28.53} & 22.04 & \underline{28.07} & 19.59 & \underline{24.23} & 26.42 & 26.57 \\
OnTheFly-NVS & 24.60 & 20.95 & 19.56 & 20.71 & 19.17 & 25.12 & 24.97 \\
ARTDECO & 25.99 & \textbf{23.82} & 23.89 & \textbf{31.72} & 22.19 & \underline{27.48} & \underline{28.73} \\
Ours & \textbf{31.42} & \underline{23.72} & \textbf{32.12} & \underline{27.04} & \textbf{29.47} & \textbf{30.94} & \textbf{30.96} \\
\midrule
Method & 8b2c0938d6 & 8be0cd3817 & ad2d07fd11 & b73f5cdc41 & e814070fd4 & f1efd25854 &  \\
\midrule
MonoGS & 12.09 & 18.21 & 9.39 & 17.74 & 19.49 & 24.47 &  \\
S3PO-GS & 18.04 & 24.29 & 18.18 & 25.67 & 22.23 & 25.27 &  \\
HI-SLAM2 & 20.16 & 21.58 & 20.55 & 26.88 & 23.52 & 25.75 &  \\
OnTheFly-NVS & 18.54 & 24.01 & 18.58 & \textbf{32.42} & 21.05 & 24.77 &  \\
ARTDECO & \underline{20.77} & \underline{29.11} & \underline{23.40} & 27.28 & \underline{28.06} & \underline{28.53} &  \\
Ours & \textbf{23.44} & \textbf{29.43} & \textbf{26.48} & \underline{30.41} & \textbf{31.13} & \textbf{31.42} &  \\
\bottomrule
\end{tabular}%
}
\end{center}
\end{table}

\begin{table}[htbp]
\begin{center}
\renewcommand{\arraystretch}{1.1}
\setlength{\tabcolsep}{2.5pt}
\footnotesize
\caption{SSIM on the ScanNet++ Dataset} 
\label{tab:track_scannetpp_ssim}%
\resizebox{1\linewidth}{!}{
\begin{tabular}{c|ccccccc}
\toprule
Method & 00777c41d4 & 126d03d821 & 1cbb105c6a & 4808c4a397 & 546292a9db & 7543973e1a & 0a7cc12c0e \\
\midrule
MonoGS & 0.474 & 0.792 & 0.800 & 0.852 & 0.367 & 0.823 & 0.813 \\
S3PO-GS & 0.759 & 0.817 & 0.847 & 0.854 & 0.671 & 0.853 & 0.818 \\
HI-SLAM2 & 0.608 & 0.870 & 0.804 & 0.869 & 0.747 & \underline{0.893} & 0.831 \\
OnTheFly-NVS & 0.655 & 0.779 & 0.781 & 0.823 & \textbf{0.847} & 0.812 & 0.835 \\
ARTDECO & \textbf{0.807} & \underline{0.904} & \underline{0.874} & \underline{0.915} & \underline{0.776} & 0.873 & \underline{0.848} \\
Ours & \textbf{0.807} & \textbf{0.907} & \textbf{0.904} & \textbf{0.931} & \underline{0.776} & \textbf{0.927} & \textbf{0.929} \\
\midrule
Method & 0d8ead0038 & 1a8e0d78c0 & 21532e059d & 251443268c & 3f15a9266d & 617dc40bca & 7079b59642 \\
\midrule
MonoGS & 0.879 & 0.505 & 0.819 & 0.708 & 0.784 & 0.816 & 0.752 \\
S3PO-GS & 0.883 & 0.769 & 0.762 & 0.835 & 0.796 & 0.817 & 0.788 \\
HI-SLAM2 & 0.915 & 0.797 & 0.886 & 0.695 & 0.833 & 0.804 & 0.800 \\
OnTheFly-NVS & 0.881 & \underline{0.815} & 0.790 & 0.802 & 0.743 & 0.818 & 0.791 \\
ARTDECO & \textbf{0.939} & 0.804 & \underline{0.915} & \underline{0.850} & \underline{0.881} & \underline{0.863} & \underline{0.833} \\
Ours & \underline{0.937} & \textbf{0.835} & \textbf{0.934} & \textbf{0.870} & \textbf{0.908} & \textbf{0.904} & \textbf{0.882} \\
\midrule
Method & 8b2c0938d6 & 8be0cd3817 & ad2d07fd11 & b73f5cdc41 & e814070fd4 & f1efd25854 &  \\
\midrule
MonoGS & 0.676 & 0.797 & 0.523 & 0.818 & 0.822 & 0.818 &  \\
S3PO-GS & 0.752 & 0.853 & 0.767 & 0.878 & 0.845 & 0.817 &  \\
HI-SLAM2 & 0.737 & 0.813 & 0.775 & 0.887 & 0.855 & 0.799 &  \\
OnTheFly-NVS & 0.751 & 0.864 & 0.773 & \textbf{0.948} & 0.841 & 0.821 &  \\
ARTDECO & \underline{0.784} & \textbf{0.913} & \underline{0.828} & 0.899 & \underline{0.909} & \underline{0.861} &  \\
Ours & \textbf{0.830} & \underline{0.910} & \textbf{0.878} & \underline{0.930} & \textbf{0.936} & \textbf{0.910} &  \\
\bottomrule
\end{tabular}%
}
\end{center}
\end{table}

\begin{table}[htbp]
\begin{center}
\renewcommand{\arraystretch}{1.1}
\setlength{\tabcolsep}{2.5pt}
\footnotesize
\caption{LPIPS on the ScanNet++ Dataset} 
\label{tab:track_scannetpp_lpips}%
\resizebox{1\linewidth}{!}{
\begin{tabular}{c|ccccccc}
\toprule
Method & 00777c41d4 & 126d03d821 & 1cbb105c6a & 4808c4a397 & 546292a9db & 7543973e1a & 0a7cc12c0e \\
\midrule
MonoGS & 0.828 & 0.458 & 0.476 & 0.396 & 0.733 & 0.490 & 0.491 \\
S3PO-GS & 0.532 & 0.382 & 0.388 & 0.371 & 0.547 & 0.402 & 0.437 \\
HI-SLAM2 & 0.582 & 0.194 & 0.331 & 0.292 & 0.370 & 0.301 & 0.329 \\
OnTheFly-NVS & 0.419 & 0.365 & 0.376 & 0.330 & \textbf{0.214} & 0.343 & 0.302 \\
ARTDECO & \underline{0.256} & \underline{0.185} & \underline{0.237} & \underline{0.175} & 0.309 & \underline{0.254} & \underline{0.277} \\
Ours & \textbf{0.244} & \textbf{0.176} & \textbf{0.183} & \textbf{0.150} & \underline{0.281} & \textbf{0.174} & \textbf{0.147} \\
\midrule
Method & 0d8ead0038 & 1a8e0d78c0 & 21532e059d & 251443268c & 3f15a9266d & 617dc40bca & 7079b59642 \\
\midrule
MonoGS & 0.503 & 0.740 & 0.475 & 0.635 & 0.480 & 0.624 & 0.683 \\
S3PO-GS & 0.486 & 0.492 & 0.472 & 0.382 & 0.437 & 0.619 & 0.532 \\
HI-SLAM2 & 0.231 & 0.335 & \textbf{0.160} & 0.513 & 0.235 & 0.431 & 0.379 \\
OnTheFly-NVS & 0.341 & \underline{0.293} & 0.396 & 0.380 & 0.420 & 0.355 & 0.391 \\
ARTDECO & \underline{0.198} & 0.314 & 0.205 & \underline{0.239} & \underline{0.216} & \underline{0.245} & \underline{0.313} \\
Ours & \textbf{0.194} & \textbf{0.268} & \textbf{0.160} & \textbf{0.207} & \textbf{0.167} & \textbf{0.142} & \textbf{0.225} \\
\midrule
Method & 8b2c0938d6 & 8be0cd3817 & ad2d07fd11 & b73f5cdc41 & e814070fd4 & f1efd25854 &  \\
\midrule
MonoGS & 0.708 & 0.531 & 0.745 & 0.541 & 0.496 & 0.615 &  \\
S3PO-GS & 0.537 & 0.344 & 0.525 & 0.304 & 0.420 & 0.607 &  \\
HI-SLAM2 & 0.466 & 0.383 & 0.398 & 0.192 & 0.284 & 0.429 &  \\
OnTheFly-NVS & 0.378 & 0.269 & 0.370 & \textbf{0.117} & 0.328 & 0.336 &  \\
ARTDECO & \underline{0.341} & \underline{0.196} & \underline{0.282} & 0.205 & \underline{0.215} & \underline{0.249} &  \\
Ours & \textbf{0.259} & \textbf{0.187} & \textbf{0.205} & \underline{0.140} & \textbf{0.150} & \textbf{0.132} &  \\
\bottomrule
\end{tabular}%
}
\end{center}
\end{table}

\begin{table}[htbp]
\begin{center}
\renewcommand{\arraystretch}{1.1}
\setlength{\tabcolsep}{5.5pt}
\footnotesize
\caption{PSNR on the ScanNetV2 Dataset} 
\label{tab:scannetv2_psnr}%
\resizebox{1\linewidth}{!}{
\begin{tabular}{c|ccccc}
\toprule
Method & scene0081\_00 & scene0354\_00 & scene0658\_00 & scene0671\_00 & scene0441\_00 \\
\midrule
MonoGS & 23.06 & 19.15 & \underline{21.68} & 22.25 & \underline{21.86} \\
S3PO-GS & 27.58 & 20.47 & \textbf{22.63} & 23.01 & \textbf{22.13} \\
HI-SLAM2 & 28.03 & \underline{23.41} & 20.79 & 25.51 & 21.81 \\
OnTheFly-NVS & 28.58 & 21.75 & 19.27 & 26.63 & 18.96 \\
ARTDECO & \underline{33.24} & 23.14 & 20.55 & \underline{28.64} & 19.65 \\
Ours & \textbf{33.44} & \textbf{30.87} & 20.18 & \textbf{29.73} & 19.33 \\
\midrule
Method & scene0461\_00 & scene0423\_00 & scene0144\_00 & scene0559\_00 & scene0316\_00 \\
\midrule
MonoGS & 22.25 & 21.72 & 21.79 & 25.86 & 19.17 \\
S3PO-GS & \textbf{24.04} & 25.52 & 24.62 & 30.00 & \underline{20.50} \\
HI-SLAM2 & \underline{22.76} & 25.19 & 28.11 & 24.02 & 17.90 \\
OnTheFly-NVS & 20.10 & 26.30 & 23.07 & 30.81 & 18.30 \\
ARTDECO & 21.09 & \underline{28.15} & \textbf{32.44} & \textbf{33.17} & 20.23 \\
Ours & 20.68 & \textbf{31.89} & \underline{31.53} & \underline{32.39} & \textbf{20.73} \\
\bottomrule
\end{tabular}%
}
\end{center}
\end{table}

\begin{table}[htbp]
\begin{center}
\renewcommand{\arraystretch}{1.1}
\setlength{\tabcolsep}{5.5pt}
\footnotesize
\caption{SSIM on the ScanNetV2 Dataset} 
\label{tab:scannetv2_ssim}%
\resizebox{1\linewidth}{!}{
\begin{tabular}{c|ccccc}
\toprule
Method & scene0081\_00 & scene0354\_00 & scene0658\_00 & scene0671\_00 & scene0441\_00 \\
\midrule
MonoGS & 0.831 & 0.837 & 0.838 & 0.827 & 0.860 \\
S3PO-GS & 0.861 & 0.848 & 0.853 & 0.838 & 0.865 \\
HI-SLAM2 & 0.845 & 0.872 & 0.807 & 0.837 & 0.845 \\
OnTheFly-NVS & 0.871 & 0.868 & 0.858 & 0.877 & 0.881 \\
ARTDECO & \textbf{0.901} & \underline{0.890} & \textbf{0.903} & \textbf{0.906} & \textbf{0.904} \\
Ours & \underline{0.896} & \textbf{0.941} & \underline{0.895} & \underline{0.905} & \underline{0.902} \\
\midrule
Method & scene0461\_00 & scene0423\_00 & scene0144\_00 & scene0559\_00 & scene0316\_00 \\
\midrule
MonoGS & 0.852 & 0.762 & 0.801 & 0.867 & 0.797 \\
S3PO-GS & 0.870 & 0.804 & 0.843 & 0.896 & 0.821 \\
HI-SLAM2 & 0.823 & 0.760 & 0.897 & 0.791 & 0.724 \\
OnTheFly-NVS & 0.870 & 0.811 & 0.828 & 0.911 & 0.830 \\
ARTDECO & \textbf{0.908} & \underline{0.845} & \textbf{0.943} & \textbf{0.936} & \underline{0.885} \\
Ours & \underline{0.896} & \textbf{0.857} & \underline{0.931} & \underline{0.926} & \textbf{0.893} \\
\bottomrule
\end{tabular}%
}
\end{center}
\end{table}

\begin{table}[htbp]
\begin{center}
\renewcommand{\arraystretch}{1}
\setlength{\tabcolsep}{5.5pt}
\footnotesize
\caption{LPIPS on the ScanNetV2 Dataset} 
\label{tab:scannetv2_lpips}%
\resizebox{1\linewidth}{!}{
\begin{tabular}{c|ccccc}
\toprule
Method & scene0081\_00 & scene0354\_00 & scene0658\_00 & scene0671\_00 & scene0441\_00 \\
\midrule
MonoGS & 0.513 & 0.449 & 0.519 & 0.484 & 0.529 \\
S3PO-GS & 0.429 & 0.399 & 0.457 & 0.443 & 0.499 \\
HI-SLAM2 & \textbf{0.199} & \underline{0.266} & 0.340 & \textbf{0.230} & \underline{0.289} \\
OnTheFly-NVS & 0.350 & 0.309 & 0.381 & 0.311 & 0.346 \\
ARTDECO & \underline{0.286} & 0.278 & \textbf{0.306} & \underline{0.262} & 0.300 \\
Ours & 0.304 & \textbf{0.174} & \underline{0.320} & \underline{0.262} & \textbf{0.287} \\
\midrule
Method & scene0461\_00 & scene0423\_00 & scene0144\_00 & scene0559\_00 & scene0316\_00 \\
\midrule
MonoGS & 0.505 & 0.620 & 0.448 & 0.495 & 0.523 \\
S3PO-GS & 0.453 & 0.541 & 0.317 & 0.392 & 0.458 \\
HI-SLAM2 & 0.397 & \underline{0.335} & \underline{0.174} & 0.370 & 0.355 \\
OnTheFly-NVS & 0.385 & 0.386 & 0.349 & 0.281 & 0.378 \\
ARTDECO & \textbf{0.304} & 0.341 & \textbf{0.163} & \textbf{0.242} & \underline{0.295} \\
Ours & \underline{0.330} & \textbf{0.299} & 0.193 & \underline{0.262} & \textbf{0.289} \\
\bottomrule
\end{tabular}%
}
\end{center}
\end{table}

\begin{table}[htbp]
\begin{center}
\renewcommand{\arraystretch}{1}
\setlength{\tabcolsep}{7pt}
\footnotesize
\caption{PSNR on the FAST-LIVO2 Dataset} 
\label{tab:FAST-LIVO2_psnr}%
\resizebox{1\linewidth}{!}{
\begin{tabular}{c|ccccc}
\toprule
Method & CBD\_Building\_01 & HKU\_Campus & Red\_Sculpture & Retail\_Street & SYSU \\
\midrule
MonoGS & 19.38 & 20.09 & 15.40 & 18.34 & 18.63 \\
S3PO-GS & 20.83 & 22.95 & 18.13 & 21.74 & 19.48 \\
HI-SLAM2 & 21.75 & 22.29 & 17.36 & \textbf{27.11} & \underline{23.49} \\
OnTheFly-NVS & 18.49 & 22.71 & 17.60 & 17.66 & 19.66 \\
ARTDECO & \textbf{26.55} & \underline{25.64} & \underline{21.09} & \underline{26.90} & 23.35 \\
Ours & \underline{25.09} & \textbf{26.84} & \textbf{22.75} & 25.98 & \textbf{26.72} \\
\bottomrule
\end{tabular}%
}
\end{center}
\end{table}

\begin{table}[htbp]
\begin{center}
\renewcommand{\arraystretch}{1}
\setlength{\tabcolsep}{7pt}
\footnotesize
\caption{SSIM on the FAST-LIVO2 Dataset} 
\label{tab:FAST-LIVO2_ssim}%
\resizebox{1\linewidth}{!}{
\begin{tabular}{c|ccccc}
\toprule
Method & CBD\_Building\_01 & HKU\_Campus & Red\_Sculpture & Retail\_Street & SYSU \\
\midrule
MonoGS & 0.683 & 0.586 & 0.520 & 0.562 & 0.595 \\
S3PO-GS & 0.720 & 0.634 & 0.571 & 0.683 & 0.618 \\
HI-SLAM2 & 0.794 & 0.662 & 0.594 & \textbf{0.897} & 0.690 \\
OnTheFly-NVS & 0.685 & 0.649 & 0.589 & 0.552 & 0.629 \\
ARTDECO & \textbf{0.861} & \underline{0.750} & \underline{0.702} & \underline{0.848} & \underline{0.734} \\
Ours & \underline{0.818} & \textbf{0.786} & \textbf{0.756} & 0.826 & \textbf{0.810} \\
\bottomrule
\end{tabular}%
}
\end{center}
\end{table}

\begin{table}[htbp]
\begin{center}
\renewcommand{\arraystretch}{1}
\setlength{\tabcolsep}{7pt}
\footnotesize
\caption{LPIPS on the FAST-LIVO2 Dataset} 
\label{tab:FAST-LIVO2_lpips}%
\resizebox{1\linewidth}{!}{
\begin{tabular}{c|ccccc}
\toprule
Method & CBD\_Building\_01 & HKU\_Campus & Red\_Sculpture & Retail\_Street & SYSU \\
\midrule
MonoGS & 0.645 & 0.759 & 0.770 & 0.720 & 0.698 \\
S3PO-GS & 0.413 & 0.550 & 0.551 & 0.410 & 0.571 \\
HI-SLAM2 & \underline{0.275} & 0.493 & 0.573 & \textbf{0.084} & 0.346 \\
OnTheFly-NVS & 0.470 & 0.427 & 0.461 & 0.517 & 0.476 \\
ARTDECO & \textbf{0.246} & \underline{0.339} & \underline{0.364} & \underline{0.200} & \underline{0.319} \\
Ours & 0.347 & \textbf{0.302} & \textbf{0.321} & 0.214 & \textbf{0.281} \\
\bottomrule
\end{tabular}%
}
\end{center}
\end{table}

\begin{table}[htbp]
\begin{center}
\renewcommand{\arraystretch}{1.1}
\setlength{\tabcolsep}{8pt}
\footnotesize
\caption{PSNR on the KITTI Dataset} 
\label{tab:KITTI_psnr}%
\resizebox{1\linewidth}{!}{
\begin{tabular}{c|cccccccc}
\toprule
Method & 00 & 02 & 03 & 05 & 06 & 07 & 08 & 10 \\
\midrule
MonoGS & 15.06 & 13.37 & 17.80 & 15.98 & 15.54 & 11.21 & 13.04 & 14.78 \\
S3PO-GS & 18.55 & 18.55 & 19.93 & 19.12 & 17.24 & 18.99 & 18.65 & 16.57 \\
HI-SLAM2 & 18.97 & \textbf{23.58} & \textbf{23.35} & 21.09 & 18.64 & 21.06 & 19.80 & \textbf{21.24} \\
OnTheFly-NVS & 15.11 & 16.65 & 18.72 & 18.33 & 16.77 & 17.56 & 18.57 & 15.01 \\
ARTDECO & \underline{21.29} & \underline{22.69} & 22.16 & \underline{22.30} & \underline{21.39} & \underline{21.14} & \underline{20.91} & 19.86 \\
Ours & \textbf{22.82} & 22.29 & \underline{23.26} & \textbf{23.59} & \textbf{23.48} & \textbf{22.35} & \textbf{21.86} & \underline{20.06} \\
\bottomrule
\end{tabular}%
}
\end{center}
\end{table}

\begin{table}[htbp]
\begin{center}
\renewcommand{\arraystretch}{1.1}
\setlength{\tabcolsep}{8pt}
\footnotesize
\caption{SSIM on the KITTI Dataset} 
\label{tab:KITTI_ssim}%
\resizebox{1\linewidth}{!}{
\begin{tabular}{c|cccccccc}
\toprule
Method & 00 & 02 & 03 & 05 & 06 & 07 & 08 & 10 \\
\midrule
MonoGS & 0.537 & 0.435 & 0.496 & 0.494 & 0.553 & 0.436 & 0.472 & 0.499 \\
S3PO-GS & 0.669 & 0.668 & 0.560 & 0.605 & 0.562 & 0.669 & 0.618 & 0.523 \\
HI-SLAM2 & 0.659 & \textbf{0.750} & \textbf{0.719} & 0.659 & 0.650 & 0.723 & 0.681 & \textbf{0.681} \\
OnTheFly-NVS & 0.576 & 0.523 & 0.563 & 0.620 & 0.564 & 0.641 & 0.675 & 0.498 \\
ARTDECO & \underline{0.759} & \underline{0.740} & 0.660 & \underline{0.731} & \underline{0.685} & \underline{0.745} & \underline{0.709} & 0.642 \\
Ours & \textbf{0.803} & 0.681 & \underline{0.676} & \textbf{0.755} & \textbf{0.726} & \textbf{0.764} & \textbf{0.745} & \underline{0.666} \\
\bottomrule
\end{tabular}%
}
\end{center}
\end{table}

\begin{table}[htbp]
\begin{center}
\renewcommand{\arraystretch}{1.1}
\setlength{\tabcolsep}{8pt}
\footnotesize
\caption{LPIPS on the KITTI Dataset} 
\label{tab:KITTI_lpips}%
\resizebox{1\linewidth}{!}{
\begin{tabular}{c|cccccccc}
\toprule
Method & 00 & 02 & 03 & 05 & 06 & 07 & 08 & 10 \\
\midrule
MonoGS & 0.719 & 0.781 & 0.714 & 0.743 & 0.731 & 0.796 & 0.832 & 0.770 \\
S3PO-GS & 0.323 & 0.326 & 0.487 & 0.414 & 0.491 & 0.383 & 0.376 & 0.530 \\
HI-SLAM2 & 0.359 & \textbf{0.248} & \textbf{0.301} & 0.376 & 0.442 & \underline{0.270} & \underline{0.310} & 0.337 \\
OnTheFly-NVS & 0.458 & 0.480 & 0.431 & 0.411 & 0.479 & 0.371 & 0.333 & 0.466 \\
ARTDECO & \underline{0.292} & \underline{0.263} & 0.341 & \textbf{0.247} & \underline{0.381} & \textbf{0.252} & 0.333 & \textbf{0.299} \\
Ours & \textbf{0.223} & 0.321 & \underline{0.303} & \underline{0.259} & \textbf{0.314} & 0.280 & \textbf{0.285} & \underline{0.318} \\
\bottomrule
\end{tabular}%
}
\end{center}
\end{table}

\begin{table}[htbp]
\begin{center}
\renewcommand{\arraystretch}{1.1}
\setlength{\tabcolsep}{5pt}
\footnotesize
\caption{PSNR on the Waymo Dataset} 
\label{tab:waymo_psnr}%
\resizebox{1\linewidth}{!}{
\begin{tabular}{c|ccccccccc}
\toprule
Method & 100613 & 106762 & 132384 & 13476 & 152706 & 153495 & 158686 & 163453 & 405841 \\
\midrule
MonoGS & 20.07 & 21.32 & 23.52 & 19.45 & 21.62 & 13.85 & 19.66 & 19.46 & 16.46 \\
S3PO-GS & 23.86 & 27.16 & 26.47 & 23.82 & 26.08 & 25.65 & 24.15 & 23.05 & 25.53 \\
HI-SLAM2 & 25.79 & 27.35 & 24.76 & 25.00 & 25.63 & 24.33 & 22.42 & 18.65 & 18.60 \\
OnTheFly-NVS & \underline{29.84} & \underline{31.23} & 26.79 & 25.36 & 27.48 & \textbf{29.39} & 24.68 & 23.71 & 23.20 \\
ARTDECO & 29.16 & \textbf{31.92} & \textbf{31.10} & \underline{26.54} & \textbf{31.57} & 27.38 & \underline{26.93} & \underline{24.43} & \underline{28.62} \\
Ours & \textbf{30.21} & 27.49 & \underline{30.35} & \textbf{28.91} & \underline{30.97} & \underline{27.55} & \textbf{28.63} & \textbf{26.62} & \textbf{29.75} \\
\bottomrule
\end{tabular}%
}
\end{center}
\end{table}

\begin{table}[htbp]
\begin{center}
\renewcommand{\arraystretch}{1.1}
\setlength{\tabcolsep}{5pt}
\footnotesize
\caption{SSIM on the Waymo Dataset} 
\label{tab:waymo_ssim}%
\resizebox{1\linewidth}{!}{
\begin{tabular}{c|ccccccccc}
\toprule
Method & 100613 & 106762 & 132384 & 13476 & 152706 & 153495 & 158686 & 163453 & 405841 \\
\midrule
MonoGS & 0.745 & 0.808 & 0.851 & 0.698 & 0.786 & 0.661 & 0.700 & 0.739 & 0.690 \\
S3PO-GS & 0.794 & 0.852 & 0.874 & 0.749 & 0.816 & 0.818 & 0.768 & 0.779 & 0.827 \\
HI-SLAM2 & 0.811 & 0.880 & 0.881 & 0.742 & 0.791 & 0.797 & 0.743 & 0.615 & 0.624 \\
OnTheFly-NVS & \textbf{0.902} & \underline{0.915} & 0.893 & \underline{0.812} & 0.864 & \textbf{0.902} & 0.803 & 0.803 & 0.784 \\
ARTDECO & \underline{0.890} & \textbf{0.923} & \textbf{0.920} & 0.779 & \textbf{0.896} & 0.862 & \underline{0.835} & \underline{0.810} & \underline{0.870} \\
Ours & 0.888 & 0.886 & \underline{0.915} & \textbf{0.863} & \underline{0.885} & \underline{0.867} & \textbf{0.864} & \textbf{0.865} & \textbf{0.889} \\
\bottomrule
\end{tabular}%
}
\end{center}
\end{table}

\begin{table}[htbp]
\begin{center}
\renewcommand{\arraystretch}{1.1}
\setlength{\tabcolsep}{5pt}
\footnotesize
\caption{LPIPS on the Waymo Dataset} 
\label{tab:waymo_lpips}%
\resizebox{1\linewidth}{!}{
\begin{tabular}{c|ccccccccc}
\toprule
Method & 100613 & 106762 & 132384 & 13476 & 152706 & 153495 & 158686 & 163453 & 405841 \\
\midrule
MonoGS & 0.623 & 0.539 & 0.456 & 0.754 & 0.667 & 0.728 & 0.648 & 0.670 & 0.638 \\
S3PO-GS & 0.445 & 0.368 & 0.327 & 0.536 & 0.505 & 0.479 & 0.429 & 0.449 & 0.392 \\
HI-SLAM2 & \textbf{0.208} & \textbf{0.162} & \textbf{0.157} & 0.319 & 0.307 & 0.298 & 0.295 & 0.403 & 0.435 \\
OnTheFly-NVS & 0.243 & 0.248 & 0.301 & \underline{0.304} & \underline{0.294} & \textbf{0.259} & 0.333 & 0.332 & 0.384 \\
ARTDECO & \underline{0.234} & \underline{0.203} & \underline{0.221} & 0.334 & \textbf{0.216} & \underline{0.275} & \textbf{0.254} & \textbf{0.283} & \underline{0.250} \\
Ours & 0.266 & 0.305 & 0.262 & \textbf{0.264} & 0.295 & 0.306 & \underline{0.256} & \underline{0.290} & \textbf{0.245} \\
\bottomrule
\end{tabular}%
}
\end{center}
\end{table}

\begin{table}[htbp]
\begin{center}
\renewcommand{\arraystretch}{1.1}
\setlength{\tabcolsep}{8pt}
\footnotesize
\caption{PSNR on the VRNeRF Dataset} 
\label{tab:vrnerf_psnr}%
\resizebox{1\linewidth}{!}{
\begin{tabular}{c|cccc}
\toprule
Method & appartment-26-2 & kitchen-26-1 & kitchen-26-2 & kitchen-26-3 \\
\midrule
MonoGS & 18.30 & 15.93 & 10.70 & 17.41 \\
S3PO-GS & 25.16 & 24.58 & 24.91 & 20.70 \\
HI-SLAM2 & 27.65 & 29.27 & 28.79 & 26.71 \\
OnTheFly-NVS & 30.87 & \textbf{32.23} & 28.79 & 25.59 \\
ARTDECO & \underline{31.98} & \underline{31.51} & \textbf{31.37} & \textbf{28.86} \\
Ours & \textbf{32.26} & 28.60 & \underline{29.94} & \underline{28.29} \\
\midrule
Method & table\_6-1 & workspace\_6\_1 & workspace\_6\_2 & workspace\_6\_4 \\
\midrule
MonoGS & 13.84 & 14.84 & 11.55 & 14.96 \\
S3PO-GS & 20.14 & 22.77 & 17.58 & 20.92 \\
HI-SLAM2 & 30.80 & \textbf{28.45} & 27.35 & 27.00 \\
OnTheFly-NVS & 28.57 & 22.97 & \underline{27.78} & \textbf{29.04} \\
ARTDECO & \underline{31.15} & 26.62 & 25.29 & 26.41 \\
Ours & \textbf{33.80} & \underline{27.67} & \textbf{28.28} & \underline{28.28} \\
\bottomrule
\end{tabular}%
}
\end{center}
\end{table}

\begin{table}[htbp]
\begin{center}
\renewcommand{\arraystretch}{1.1}
\setlength{\tabcolsep}{8pt}
\footnotesize
\caption{SSIM on the VRNeRF Dataset} 
\label{tab:vrnerf_ssim}%
\resizebox{1\linewidth}{!}{
\begin{tabular}{c|cccc}
\toprule
Method & appartment-26-2 & kitchen-26-1 & kitchen-26-2 & kitchen-26-3 \\
\midrule
MonoGS & 0.678 & 0.539 & 0.509 & 0.527 \\
S3PO-GS & 0.834 & 0.807 & 0.834 & 0.680 \\
HI-SLAM2 & 0.866 & 0.881 & 0.908 & 0.852 \\
OnTheFly-NVS & 0.912 & \textbf{0.922} & \underline{0.921} & 0.857 \\
ARTDECO & \underline{0.915} & \underline{0.912} & \textbf{0.937} & \textbf{0.895} \\
Ours & \textbf{0.916} & 0.879 & 0.915 & \underline{0.871} \\
\midrule
Method & table\_6-1 & workspace\_6\_1 & workspace\_6\_2 & workspace\_6\_4 \\
\midrule
MonoGS & 0.619 & 0.578 & 0.466 & 0.606 \\
S3PO-GS & 0.755 & 0.774 & 0.694 & 0.750 \\
HI-SLAM2 & 0.908 & \textbf{0.860} & 0.840 & 0.826 \\
OnTheFly-NVS & 0.898 & 0.824 & \textbf{0.873} & \textbf{0.901} \\
ARTDECO & \underline{0.913} & 0.831 & 0.816 & 0.841 \\
Ours & \textbf{0.947} & \underline{0.860} & \underline{0.863} & \underline{0.875} \\
\bottomrule
\end{tabular}%
}
\end{center}
\end{table}

\begin{table}[htbp]
\begin{center}
\renewcommand{\arraystretch}{1.1}
\setlength{\tabcolsep}{8pt}
\footnotesize
\caption{LPIPS on the VRNeRF Dataset} 
\label{tab:vrnerf_lpips}%
\resizebox{1\linewidth}{!}{
\begin{tabular}{c|cccc}
\toprule
Method & appartment-26-2 & kitchen-26-1 & kitchen-26-2 & kitchen-26-3 \\
\midrule
MonoGS & 0.625 & 0.711 & 0.744 & 0.658 \\
S3PO-GS & 0.441 & 0.468 & 0.414 & 0.557 \\
HI-SLAM2 & 0.350 & 0.269 & 0.220 & 0.309 \\
OnTheFly-NVS & 0.307 & \underline{0.225} & 0.223 & 0.246 \\
ARTDECO & \underline{0.232} & \textbf{0.205} & \textbf{0.184} & \textbf{0.200} \\
Ours & \textbf{0.192} & 0.270 & \underline{0.189} & \underline{0.206} \\
\midrule
Method & table\_6-1 & workspace\_6\_1 & workspace\_6\_2 & workspace\_6\_4 \\
\midrule
MonoGS & 0.744 & 0.815 & 0.774 & 0.753 \\
S3PO-GS & 0.528 & 0.515 & 0.719 & 0.577 \\
HI-SLAM2 & \underline{0.219} & \underline{0.249} & 0.281 & 0.318 \\
OnTheFly-NVS & 0.245 & 0.300 & \underline{0.251} & \textbf{0.256} \\
ARTDECO & \textbf{0.200} & 0.261 & 0.294 & 0.274 \\
Ours & 0.230 & \textbf{0.218} & \textbf{0.221} & \underline{0.261} \\
\bottomrule
\end{tabular}%
}
\end{center}
\end{table}


%
%

\end{document}